\documentclass[twoside,11pt]{article}

\usepackage[abbrvbib]{jmlr2e}

\usepackage{amsmath}
\usepackage{mathtools} 
\usepackage{amsfonts}
\usepackage{amssymb}
\usepackage{bbm}
\usepackage{algorithm}
\usepackage{algorithmic}

\DeclareMathOperator*{\argmin}{argmin} 
\usepackage{url}
\usepackage{booktabs}
\usepackage[export]{adjustbox}
\usepackage{caption}
\usepackage{varwidth}
\newsavebox\tmpbox

\usepackage{lastpage}
\jmlrheading{24}{2023}{1-\pageref{LastPage}}{3/21; Revised 5/22}{1/23}{21-0326}{Jon Vadillo, Roberto Santana and Jose A. Lozano}
\ShortHeadings{title}{Vadillo, Santana and Lozano}

\ShortHeadings{Extending Adv. Attacks to Produce Adv. Class Probability Distributions}{Vadillo, Santana and Lozano}
\firstpageno{1}

\usepackage{threeparttable}
\usepackage{siunitx}

\usepackage{xcolor}

\begin{document}

\title{Extending Adversarial Attacks to Produce Adversarial Class Probability Distributions}

\author{\name Jon Vadillo \email jon.vadillo@ehu.eus \\
       \addr Department of Computer Science and Artificial Intelligence\\
       University of the Basque Country UPV/EHU\\
       20018 Donostia, Spain.
       \AND
       \name Roberto Santana \email roberto.santana@ehu.eus \\
       \addr Department of Computer Science and Artificial Intelligence\\
       University of the Basque Country UPV/EHU\\
       20018 Donostia, Spain.
       \AND
       \name Jose A. Lozano \email ja.lozano@ehu.eus \\
       \addr Department of Computer Science and Artificial Intelligence\\
       University of the Basque Country UPV/EHU\\
       20018 Donostia, Spain.\\
       \addr Basque Center for Applied Mathematics (BCAM)\\
       48009 Bilbao, Spain.}

\editor{Sathiya Keerthi}

\maketitle

\begin{abstract}
Despite the remarkable performance and generalization levels of deep learning models in a wide range of artificial intelligence tasks, it has been demonstrated that these models can be easily fooled by the addition of imperceptible yet malicious perturbations to natural inputs. These altered inputs are known in the literature as adversarial examples. In this paper, we propose a novel probabilistic framework to generalize and extend adversarial attacks in order to produce a desired probability distribution for the classes when we apply the attack method to a large number of inputs. This novel attack paradigm provides the adversary with greater control over the target model, thereby exposing, in a wide range of scenarios, threats against deep learning models that cannot be conducted by the conventional paradigms. We introduce four different strategies to efficiently generate such attacks, and illustrate our approach by extending multiple adversarial attack algorithms. We also experimentally validate our approach for the spoken command classification task and the Tweet emotion classification task, two exemplary machine learning problems in the audio and text domain, respectively. Our results demonstrate that we can closely approximate any probability distribution for the classes while maintaining a high fooling rate and even prevent the attacks from being detected by label-shift detection methods.

\end{abstract}


\section{Introduction}
\label{sec:introduction}

Deep Neural Networks (DNNs) are currently the core of a wide range of technologies applied in critical tasks, such as self-driving vehicles \citep{xu2017end, muller2006off}, identity recognition systems \citep{boles2017voice, sun2015deepid3, masi2018deep} or malware detection \citep{huang2016mtnet, saxe2015deep}, and effectiveness and robustness are therefore two fundamental requirements for these models. However, it has been found that DNNs can be easily deceived by inputs perturbed imperceptibly for humans, known as adversarial examples \citep{szegedy2013intriguing}, which implies a security breach that can be maliciously exploited by an adversary.  Although the study of this phenomenon has focused mainly on computer vision tasks, it has been shown that adversarial examples can be found in multiple tasks and domains, such as audio \citep{kereliuk2015deep,carlini2018audio,alzantot2018did} or natural language processing \citep{yang2020greedy, samanta2017towards, alzantot2018generating}.

Most adversarial perturbation generation methods can be taxonomized in different groups according to the scope of the objective of the adversarial attack. The most frequent methods focus on fooling a target model during its prediction phase, with varying degrees of generality, such as individual perturbations (designed for one particular input), single-class perturbations \citep{vadillo2019universal,gupta2019method,zhang2020cduap} (designed to fool any input of one particular class) or universal (input-agnostic) perturbations \citep{moosavi2017universal}. These perturbations can be created with the objective of changing the (originally correct) output to any other (incorrect) class, called \textit{untargeted attacks}, or even forcing the model to produce a particular target class. In the last case, we refer to them as \textit{targeted attacks}, which provide greater control over the target model than untargeted attacks.

Overall, the literature on adversarial attacks has mainly focused on ``single-instance'' scenarios, where the goal is to minimally manipulate the input at hand, so that the model misclassifies that instance in particular. Only a few works have considered ``multiple-instance'' scenarios, where adversarial attacks are used to achieve malicious goals that can only be realized by considering multiple inputs (e.g., by generating multiple attacks).

In \cite{lin2017tactics,tretschk2018sequential,hussenot2020copycat}, adversarial examples are sequentially created and fed to reinforcement learning models in order to control their behavior in the long run. More particularly, a sequence of adversarial examples is generated in \cite{lin2017tactics} to force the model to take a preferred sequence of actions, which can be used to guide the agent towards a particular state of interest. In \cite{tretschk2018sequential}, the sequence of attacks attempts to impose an adversarial reward of interest on the victim policy at test time (i.e., leading the agent's policy to maximize the imposed adversarial reward). Similarly, the goal of imposing an adversarial target-policy is pursued in \cite{hussenot2020copycat}. 
Other works attempted to sequentially generate \textit{adversarial} inputs in order to introduce adversarial concept drifts in streaming classification scenarios \citep{sethi2018handling,kantchelian2013approaches,korycki2020adversarial}, which can lead to a drop in the performance of the model. A comprehensive taxonomy of adversarial concept drifts is proposed in \cite{korycki2020adversarial}, where different types of goals are discussed, such as injecting a sequence of \textit{corrupted} instances to make the adaptation to a real concept drift difficult, or injecting a sequence of adversarial instances which form a coherent concept and which are capable of inducing a concept drift.

\subsection{Objective}
In this work, we introduce a novel ``multiple-instance'' adversarial attack strategy. In particular, we propose a method which provides the adversary with the ability not only to deceive the model by adding imperceptible perturbations to the inputs, but also to control the frequency or proportion with which each class is predicted by the model, even in scenarios where we can only introduce very small amounts of distortion to the inputs.

Let us consider a target machine learning model $\mathcal{M}$ that implements a classification function $f: X \rightarrow Y$, where $X\subseteq \mathbb{R}^d$ represents the $d$-dimensional input space, and $Y=\left\lbrace y_1, \dots, y_k \right\rbrace$ the set of possible output classes, being $y_i$ the $i$-th class, $i=1,\dots,k$.  
The main objective of this paper is to create an attack method $\Phi$ that is able to efficiently produce adversarial examples $x'=\Phi(x)$ not only with the objective of achieving $f(x') \neq f(x)$ for every input,  but also to accomplish the objective of producing a specific probability distribution for the classes $\widetilde{\mathcal{P}}(Y)=(\tilde{p}_{1}, \dots, \tilde{p}_{k} )$  after multiple attacks, that is:
\begin{equation}
\mathbb{P}_{x\sim \mathcal{P}(X)} \left[ f(\Phi(x))\!=\!y_i \right] = \tilde{p}_{i} \ , \ 1\leq i \leq k,
\end{equation} 
where $\mathcal{P}(X)$ represents the probability distribution of the \textit{natural inputs}.

\subsection{Applications and Use Cases}
\label{sec:applications_use_cases}
The idea of controlling the probability distribution of the classes produced by the adversarial examples (after sending multiple adversarial inputs to the model) provides a novel perspective to design such attacks.
First, this attack can be used to produce drifts in the probability distribution of the classes, commonly referred to as \textit{target shift} \citep{zhang2013domain}, \textit{label shift} \citep{lipton2018detecting,garg2020unified} or as \textit{prior probability shift} \citep{quinonero-candela2009when,saerens2002adjusting,biswas2021ensuring} in the literature. 
Indeed, it has been shown that such drifts can degrade the performance of the classifiers \citep{lipton2018detecting,vucetic2001classification,saerens2002adjusting}, or imply ethical issues when such changes cause the predictions of the models to be biased or unfair \citep{vucetic2001classification,biswas2021ensuring}.
Thus, strategies have been proposed to detect such changes in the probability distribution of the classes during the prediction phase, and even to correct or ``adjust'' the decisions of the models to, accounting for those changes, improve the classification performance of the model \citep{vucetic2001classification,saerens2002adjusting,lipton2018detecting}.

Secondly, controlling the output probability of the classes might also be of particular interest for those cases in which the frequency with which each class is predicted for multiple inputs (i.e., the class distribution) is more relevant than the individual predictions given for each input. This is, for instance, the case of the quantification learning paradigm \citep{gonzalez2017why,gonzalez2017review,qi2021framework}. Representative examples and domains of this paradigm are opinion mining, sentiment analysis or collective information retrieval in social networks \citep{gao2016classification,milli2015quantification,giachanou2016it,zarrad2019realtime}, where the main focus can be, for instance, on accurately estimating the frequency of a particular opinion among a population. Other sensitive domains where the aggregated results of the output class is relevant is epidemiology \citep{king2008verbal}, for example, to estimate the cause-specific mortality in a population or the prevalence of a specific disease, which might be crucial to tackle it.
Thus, in all these applications, maliciously changing the ratios with which each class is predicted for multiple inputs might bring about critical consequences, such as biased estimations of the population opinion or an incorrect screening of the prevalence of a disease, leading to an ineffective action plan.

Finally, whereas, to the best of our knowledge, defensive methods against such ``multiple-instance'' adversarial attacks have not been proposed (since all of them focus on counteracting attacks in the ``single-instance'' scenario), recent works have shown that a label-drift might be a clear indicator of adversarial activity  \citep{rabanser2019failing}. Thus, label-drift detection methods \citep{lipton2018detecting,rabanser2019failing} could be straightforwardly applied as defensive countermeasures in order to detect that an adversary is sending multiple adversarial attacks to the models \citep{rabanser2019failing}.
Therefore, from the adversary's perspective, controlling the frequency with which each class is predicted allows, for instance, the same probability distribution produced by the target DNN on clean inputs to be replicated, making the attacks less likely to be detected by label-drift detection mechanisms.

\subsection{Contribution}
For all these reasons, maliciously controlling the probability distribution of the classes can lead to more ambitious and complex attacks. However, the current adversarial attacks proposed in the literature are not capable of controlling such distributions. The main contribution of this paper is to fill this gap by introducing a probabilistic framework with which any targeted adversarial attack can be extended to produce not only a misclassification in a DNN for the incoming inputs, but also any target probability distribution of the output classes after multiple attacks. In particular, we propose four different methods to create the optimal strategies to guide such attacks, and we validate them by extending a wide range of adversarial attacks for two exemplary machine learning problems.
The effectiveness of the proposed four strategies is compared under multiple criteria, such as the similarity of the produced probability distributions and the target distributions, the percentage of inputs fooled by the attack or the number of parameters to be optimized for each method.\footnote{Our code is publicly available at: \url{https://github.com/vadel/ACPD}.}

The rest of this paper is organized as follows. Section \ref{sec:basic_concepts} provides a detailed description of the proposed adversarial attack strategy, and specifies a number of assumptions and key concepts. Section \ref{sec:methods} introduces four different approaches to produce a target probability distribution for the output classes. Section \ref{sec:validation} describes the experimental setups used to evaluate and compare the effectiveness of the approaches introduced. This section also includes the experimental results. Section \ref{sec:exp_shift_detection} illustrates how our methods can be applied to produce label-shifts in streaming classification scenarios without alerting label-shift detection mechanisms. Section \ref{sec:conclusions} concludes the paper.

\section{Producing Specific Class Probability Distributions}
\label{sec:basic_concepts}
We focus on defining an attack approach in which the application of the attack for many incoming inputs $x$, assuming an input data distribution $x \sim \mathcal{P}(X)$, can produce not only a misclassification for every $x$, but also a desired (fixed) probability distribution of the predicted classes by the target model $\widetilde{\mathcal{P}}(Y)=(\tilde{p}_{1}, \dots, \tilde{p}_{k})$.

\subsection{Assumptions and Key Concepts}
\label{sec:assumptions}
In this section, we specify a number of assumptions and concepts that will be used to develop our methodology.

First of all, we assume that the clean input $x$ is correctly classified by the target classifier, and that $f(\Phi(x)\!=\!x')\neq  f(x)$, in order to ensure that the attack is actually fooling the model. In addition, being $\varphi: \mathbb{R}^d \times \mathbb{R}^d \rightarrow \mathbb{R}$ a distortion metric and $\epsilon$ a maximum distortion threshold, we require the adversarial example to satisfy $\varphi(x,x')\leq \epsilon$, to ensure that $x'$ is as similar as possible to $x$. In the literature, common choices for $\varphi$ are $\ell_p$ norms such as the $\ell_2$ or the $\ell_{\infty}$ norm.

The approach we introduce will use a targeted adversarial attack as a basis, that is, attacks capable of forcing the model to produce a particular target class $f(x')=y_j$. However, setting a maximum distortion supposes that we may not reach every possible target class by adversarially perturbing an input sample. For this reason, we consider that $y_j$ is a \textit{reachable class} from $x$ if it is possible to generate a targeted adversarial example $x'$, so that $\varphi(x, x')\!\leq\!\epsilon$ and $f(x')=y_j$, and it will be denoted as $\Phi(x)\rightarrow y_j$. We assume that $f(x)$ is always a reachable class. However, if there are no reachable classes $y_j \neq f(x)$, we will consider that we can not create any \textit{valid} adversarial example for $x$.

\subsection{Attack Description}
\label{sec:attack_description}
The main rationale of the approach we introduce is to \textit{guide} a targeted adversarial attack method $\Psi$ in order to achieve the global objective of producing any probability distribution of the output classes $\widetilde{\mathcal{P}}(Y)$, while maintaining a high fooling rate and minimally distorted inputs. To enable such attacks, our method consists of generalizing $\Psi$ to be stochastic, so that the target class is randomly selected, and the probability of transitioning from the class $y_i$ to the class $y_j$ depends on the source class $y_i$ and the input $x$ at hand. 

These probabilities will be represented by a transition matrix $T=[t_{i,j}]_{i,j=1}^k$, where $t_{i,j}$ represents the probability of transitioning from the class $y_i$ to the class $y_j$. In the event that, given an input $x$ of class $y_i$, it is not possible to reach all the classes without exceeding the distortion limit, the probability of transitioning to a non-reachable class will be set to zero, and the probability distribution $(t_{i,1}, \dots, t_{i,k})$ will be normalized accordingly. Thus, being $\mathcal{Y}=\{y_j : \Phi(x) \rightarrow y_j \}$ the set of reachable classes for one particular input $x$ of class $y_i$, the probability of selecting $y_j$ as the target class is determined by:
\begin{equation}
\label{eq:normalization}
t_{i,j}' = 
\begin{cases} 
       \frac{t_{i,j}}{\sum_{ y_r\in \mathcal{Y}}t_{i,r}} &  \textit{if } y_j \in \mathcal{Y} \\ 
        0  & \text{otherwise}.
\end{cases}
\end{equation} 
By modeling the decision to move from one class to another in this way, it is possible to approximate with which probability the model will predict each class. Algorithm \ref{algorithm_general_deepfool} provides the pseudocode of this approach.

\begin{algorithm}[!t]
 \caption{Generating adversarial class probability distributions.
 }
  \label{algorithm_general_deepfool}
\begin{algorithmic}[1]
\REQUIRE {A classification model $f$, a set of classes $Y=\{y_1,\dots,y_k\}$, a targeted adversarial attack method $\Psi$, a distortion metric $\varphi$, a maximum distortion threshold $\epsilon$, a transition matrix $T$, a set of inputs samples $\hat{\mathcal{X}}$.}
\FORALL{ $x\in \hat{\mathcal{X}}$}
   \STATE $reachable[1,\dots,k] \gets$ initialize with \textit{False}.
   \FOR{$j \in \{1,\dots,k\}$}
  	  \STATE $v_j \gets$ use $\Psi$ to generate an adversarial perturbation for $x$ targeting class $y_j$\\
  	  \IF{$f(x+v_j) = y_j \wedge \varphi(x,x+v_j) \leq \epsilon$}
  	  	\STATE $reachable[j] \gets True$ 
	 \ENDIF  
  \ENDFOR 
  \STATE $\mathcal{Y} \gets \{y_j\in Y : reachable[j]=True\}$
  \STATE $(t_{i,1},\dots,t_{i,k}) \gets $ probability distribution in the row of $T$ corresponding to the source class $y_i=f(x)$.
  \STATE $t_{sum} \gets \sum_{y_j \in \mathcal{Y}}(t_{i,j})$
  \IF{ $t_{sum}=0$}
      \STATE Feed $x$ to the model $f$. 
  \ELSE
      \FOR{$j \in \{1,\dots,k\}$}
      	\IF{$reachable[j]=True$}
      		\STATE $t'_{i,j} \gets \frac{t_{i,j}}{t_{sum}}$
      	\ELSE
      		\STATE $t'_{i,j} \gets 0$	
      	\ENDIF
      \ENDFOR
      \STATE $y^* \gets$ randomly select a class according to the probabilities $(t_{i,1}',\dots,t_{i,k}')$.\\
      \STATE Select the adversarial example with the targeted perturbation $v^*$ corresponding to class $y^*$:\\
      $x' \gets x+v^*$ 
      \STATE Feed $x'$ to the model $f$.
  \ENDIF
\ENDFOR
\end{algorithmic}
\end{algorithm}

Note that the probability $t_{i,i}'$ represents the probability of maintaining an input in its own class $y_i$, and, therefore, these values should be as low as possible in order to ensure that we maximize the number of inputs that will fool the model. However, depending on the probability distribution of the classes we want to produce, a nonzero value for these probabilities may be needed to achieve such goals, for instance, if we require a high probability for one class but this class is seldomly reached from inputs belonging to the rest of classes. How to obtain transition matrices $T$ that comply all the aforementioned conditions will be discussed in detail in Section \ref{sec:methods}, where four different methods are proposed.

Finally, we would like to point out that our approach is not subject to any particular targeted adversarial attack strategy, and, therefore, it is agnostic with regard to the particularities of the selected strategy  (for example, the amount of information about the model that the adversary can exploit to generate the attacks). 
This allows the adversary to select the most appropriate attack depending on the requirements of the problem or scenario. For instance, for scenarios where the computation time is a critical aspect or for problems with a large number of classes, the adversary can opt for adversarial attacks with low computational cost.\footnote{It is worth pointing out that the process of generating an adversarial example for each target class is fully parallelizable, since each targeted attack is independent of the others, making our method applicable in practice even for problems with a large number of classes.}
On the other hand, in less restrictive scenarios, the adversary can employ more effective attacks, at the expense of higher computational cost.

\section{Constructing Optimal Transition Matrices to Guide Targeted Attacks}
\label{sec:methods}
In this section we introduce different strategies to construct the optimal transition matrix $T$ which, used to stochastically decide the class transitions, produces a target probability distribution $\widetilde{\mathcal{P}}(Y)= (\tilde{p}_1,\dots,\tilde{p}_k)$ for the output classes. Formally, being $\mathcal{X}=\{x_1, \dots, x_n \}$ a set of inputs sampled from a data distribution $\mathcal{P}(X)$, and $\mathcal{P}(Y)=(p_1,\dots,p_k)$ the original probability distribution of the classes assigned by the target classifier, we want to obtain a transition matrix $T$ that satisfies:
\begin{equation}
\label{eq:py_t_pyobj}
(p_1,\dots,p_k)\!
\begin{pmatrix}
t_{1,1} & t_{1,2} & \cdots & t_{1,k} \\
t_{2,1} & t_{2,2} & \cdots & t_{2,k} \\
\vdots  & \vdots  & \ddots & \vdots  \\
t_{k,1} & t_{k,2} & \cdots & t_{k,k} 
\end{pmatrix}\! = \! (\tilde{p}_1,\dots,\tilde{p}_k).
\end{equation}
We will define the problem of finding such matrices as a linear program, and, in order to restrict the possible values of $T$, different strategies will be introduced.

The main objective is to ensure that $T$ satisfies Equation \eqref{eq:py_t_pyobj}. Therefore, this equation is added as a constraint of the linear program. The second objective is to maximize the expected fooling rate of the attack, that is, to minimize the probability of keeping the original class predicted by the model unchanged. This will be achieved by adding the sum of the diagonal of $T$ as a component of the objective function of the linear program, which will be minimized.

It is important to note that, although multiple optimal solutions may exist for these problems, it is not expected that all of them will produce the same approximation to the target probability distribution of the classes $\widetilde{P}(Y)$ when applied in the prediction phase of the classifier  (i.e., when our attacks are put in practice).
For instance, if many of the values of $T$ are zero, then these transitions can not be carried out, resulting in inaccurate approximations of $\widetilde{P}(Y)$. Similarly, if many of the transitions are not possible in practice (something that can happen for low distortion budgets or problems in which targeted attacks can not always be successfully generated), then different transition matrices could produce very different results.
For these reasons, the four methods that will be introduced in this section will rely on different strategies to increase the effectiveness of the matrices in the prediction phase.
In addition, they will differ in the amount of information they use from $\mathcal{X}$. While our first method is almost agnostic, the subsequent three use more informative approaches.

\subsection{Method 1: Agnostic Method (AM)}
\label{sec:method1}
The first method will follow an almost agnostic approach to generate the transition matrix $T$, where the only information that will be used is the initial probability distribution $\mathcal{P}(Y)$. Therefore, the results obtained with this method will be used to compare the gain that the following methods imply, in which more informed transition matrices will be created.

Thus, this method consists of directly searching for a transition matrix $T$ that satisfies Equation \eqref{eq:py_t_pyobj}, while minimizing the sum of the diagonal of $T$. To avoid a high number of null $t_{i,j}$ probabilities outside the diagonal of $T$, an auxiliary variable matrix $L=[l_{i,j}]_{i,j=1}^k$ will be introduced, so that $l_{i,i}=0$ and $0\leq l_{i,j} \leq \xi, i\neq j$, with $\xi \in \mathbb{R}$ and $\xi\ll 1/k$. Each $l_{i,j}$ will be included in the set of restrictions as a lower bound of $t_{i,j}$ to require a minimum probability, $l_{i,j} \leq t_{i,j}$, $i\neq j$. At the same time, the values in $L$ will be maximized in the  objective function of the linear program.

Taking into account all these basic requirements, the optimal transition matrix $T$ can be obtained by solving the following linear program:
\begin{equation}
\label{eq:linear_program_agnostic}
\begin{aligned}
\textrm{min} & \quad  z = \gamma_1 \cdot \sum_{i=1}^k t_{i,i} - \gamma_2 \cdot \sum_{i=1}^k\sum_{\substack{j=1 \\ j\neq i}}^k l_{i,j}\\
\textrm{s.t.} & \quad \mathcal{P}(Y)\cdot T=\widetilde{\mathcal{P}}(Y) \\
& \quad \sum_{j=1}^k t_{i,j} = 1 & \forall i\!\in\!\{1,\dots, k \} \\
& \quad t_{i,j} \geq l_{i,j} & \forall i,j \in  \{1,\dots, k \}, i\neq j\\
& \quad 0 \leq  t_{i,j} \leq 1   & \forall i,j \in  \{1,\dots, k \}\\
& \quad 0 \leq  l_{i,j} \leq \xi & \forall i,j \in  \{1,\dots, k \}.\\
\end{aligned}
\end{equation}

For the sake of generality, a coefficient $\gamma\in\mathbb{R}$ was included for each of the main terms in the objective function, allowing their importance to be traded off. This will also be done in the subsequent methods.

\subsection{Method 2:  Upper-bound Method (UBM)}
\label{sec:method2}
In the second method, we will extend the linear program introduced in the AM to include an additional restriction to the values of $T$ in order to capture more accurate information about the feasible class transitions associated to the perturbations. In this approach, an auxiliary matrix $R=[r_{i,j}]_{i,j=1}^k$ will be considered, in which $r_{i,j}$ represents the number of samples in $\mathcal{X}$ that, with a ground-truth class $y_i$, can reach the class $y_j$:
\begin{equation}
\label{eq:matrix_r}
r_{i,j} = |\{ x \! \in \! \mathcal{X} : f(x)\!=\!y_i \wedge \Phi(x)\!\rightarrow\!y_j \}|.
\end{equation}
We assume that the ground-truth class of an input is always reachable, and therefore, ${r_{i,i}=|\{x\in \mathcal{X} : f(x)=y_i\}|}$, ${1 \leq i \leq k}$. If we divide each $r_{i,j}$ by the number of inputs of class $y_i$, the value will represent the proportion of samples in $\mathcal{X}$ which, with a ground-truth class $y_i$, can reach the class $y_j$:
\begin{equation}
r'_{i,j} = \frac{r_{i,j}}{|\{ x \! \in \! \mathcal{X} : f(x)\!=\!y_i \}|}.
\end{equation}
Note that we are estimating, by using the set $\mathcal{X}$, the proportion of successful targeted attacks that it is possible to create for the inputs coming from $\mathcal{P}(X)$, for any pair of source class $y_i$ and target class $y_j$. Therefore, to generate a more informed transition matrix $T$, we will maintain the following restriction: $t_{i,j}\leq r'_{i,j}$. The aim of this restriction is to avoid assigning transition probabilities so high that, in practice, it will be unlikely to obtain them due to the distortion threshold, which may imply a loss of effectiveness regarding the global objective of producing $\widetilde{\mathcal{P}}(Y)$, as the algorithm may not be able to successfully follow the guidance of $T$. 

However, setting an upper bound to the values of $T$ according to the values of $R$ may imply increasing the values in the diagonal, decreasing the fooling rate expectation. Moreover, those restrictions can be too strict for low distortion thresholds, making it impossible to find feasible solutions in the linear program for a large number of cases (see Table \ref{tab:success_perc}). Thus, to relax these restrictions, we will consider an auxiliary set of variables $0\leq \eta_{i,j} \leq 1$, that will act as upper thresholds for the $t_{i,j}$ values in the matrix $T$, and which will be minimized in the objective function. Based on all these facts, we will generate the optimal transition matrix $T$ by solving the following linear program:
\begin{equation}
\label{eq:linear_program}
\begin{aligned}
\textrm{min} & \quad  z = \gamma_1 \cdot \sum_{i=1}^k t_{i,i} - \gamma_2 \cdot \sum_{i=1}^k\sum_{\substack{j=1 \\ j\neq i}}^k l_{i,j} + \gamma_3 \cdot \mathrlap{\sum_{i=1}^k\sum_{j=1}^k \eta_{i,j}}  \\
\textrm{s.t.} & \quad \mathcal{P}(Y)\cdot T=\widetilde{\mathcal{P}}(Y) \\
& \quad \sum_{j=1}^k t_{i,j} = 1 & \forall i\!\in\!\{1,\dots, k \} \\
& \quad t_{i,j} \geq l_{i,j} & \forall i,j \in  \{1,\dots, k \}, i\neq j\\
& \quad 0 \leq  t_{i,j} \leq 1  & \forall i,j \in  \{1,\dots, k \}\\
& \quad 0 \leq  l_{i,j} \leq \xi & \forall i,j \in  \{1,\dots, k \}\\
& \quad t_{i,j} \leq r_{i,j}' + \eta_{i,j} & \forall i,j \in  \{1,\dots, k \} \\
& \quad 0\leq \eta_{i,j} \leq 1 & \forall i,j \in  \{1,\dots, k \}. \\
\end{aligned}
\end{equation}

\subsection{Method 3:  Element-wise Transformation Method (EWTM)}
\label{sec:method3}

The main drawback of the strategy used in the UBM is that establishing bounds for every value of $T$ can significantly limit the space of possible transition matrices, reducing the range of target probability distributions $\mathcal{\widetilde{P}}(Y)$ that can be produced. Therefore, a relaxation of those restrictions is required in order to achieve feasible solutions, which at the same time could, however, reduce the effectiveness of the approach.  

In addition, even if it is estimated, using the set $\mathcal{X}$, that it is not possible to move more than a certain proportion of cases $r'_{i,j}$ from the class $y_i$ to the class $y_j$, in some cases it can be necessary to assign values higher than $r'_{i,j}$ to $t_{i,j}$, for example, to produce $y_j$ with a very high probability. In such cases, even if reaching the class $y_j$ from the class $y_i$ is unlikely, we can specify that when this transition is possible, it should be produced with a high probability.

Therefore, in order to be able to accurately approximate a wider range of distributions $\mathcal{\widetilde{P}}(Y)$, the EWTM does not impose bound constraints on the values of $T$. Apart from that, the row-normalized version of $R$ will be used in this method, denoted as $\hat{R}$, which already represents a transition matrix. In particular, the probability distribution $\mathcal{P}(Y)\hat{R}$ is the one that would be achieved if the target class of each input $x$ were uniformly selected in the set of reachable classes for $x$. As our goal is to produce $\widetilde{\mathcal{P}}(Y)$, we aim to find an auxiliary matrix $Q=[q_{i,j}]_{i,j=1}^k$, so that  $\mathcal{P}(Y) (\hat{R} \odot Q) =\widetilde{\mathcal{P}}(Y)$, where the operator $\odot$ represents the Hadamard (element-wise) product. If we denote $T=\hat{R}\odot Q$, we can generate $T$ by solving the following linear program:
\begin{equation}
\label{eq:linear_program_2_elementwise}
\begin{aligned}
\textrm{min} & \quad  z = \gamma_1 \cdot \sum_{i=1}^k \hat{r}_{i,i}\cdot q_{i,i} + \gamma_2 \cdot \eta \\
\textrm{s.t.} & \quad \mathcal{P}(Y)\cdot (\hat{R}  \odot Q) =  \widetilde{\mathcal{P}}(Y) \\
& \quad \sum_{j=1}^k \hat{r}_{i,j}\cdot q_{i,j} = 1 & \forall i\!\in\!\{1,\dots, k \} \\
& \quad 0 \leq  \hat{r}_{i,j}\cdot q_{i,j} \leq 1   & \forall i,j \in  \{1,\dots, k \}\\
& \quad 0 \leq q_{i,j} \leq \eta  & \forall i,j \in  \{1,\dots, k \}\\
& \quad 0 \leq \eta
\end{aligned}
\end{equation}
The values in $Q$ will be minimized by  including, as a decision variable, an upper bound $\eta$ for its values,  avoiding excessive transformations of the matrix $\hat{R}$ which would lead to losing its influence, and, therefore, resembling an \textit{agnostic} approach similar to the AM.

\subsection{Method 4:  Chain-rule Method (CRM)}
\label{sec:method4}
As explained in Section \ref{sec:attack_description}, even if we specify a probability $t_{i,j}$ for every possible transition, in practice, an input sample may not be able to reach any possible class without surpassing the maximum distortion allowed, so we need to normalize those probabilities to consider only the reachable classes from that input. However, the two previous methods have not considered this effect during the optimization process of the transition matrices, which may cause a reduction in the effectiveness of the resulting matrices when they are applied during the prediction phase of the model. For this reason, in this method, we will make use of that information with the aim of achieving a more informed attack.

To construct the transition matrix $T$, we start by estimating the probabilities that an input $x$ of class $y_i$ can only reach a particular subset of the classes $\mathcal{S}\subseteq Y$ (i.e., ${y_j\!\in \!\mathcal{S} \Leftrightarrow \Phi(x)\!\rightarrow\!y_j}$). We will denote these probabilities 
\begin{equation}
\label{eq:m4_Psi}
{P_{x\sim \mathcal{P}(X)}(\mathcal{S}|f(x)=y_i)},
\end{equation}
or $P(\mathcal{S}|y_i)$ for simplicity. In order to estimate these values, the set of reachable classes $\mathcal{S}$ will be computed for each $x\in \mathcal{X}$, and the frequency of each subset will be calculated. 

The next step is to define the probability that an input $x$ of class $f(x)=y_i$ and with a set of reachable classes $\mathcal{S}$ will be moved from $y_i$ to the class $y_j$, that is, 
\begin{equation}
\label{eq:m4_vij}
{P_{x\sim \mathcal{P}(X)}(y_j | f(x)=y_i, \mathcal{S})},
\end{equation}
or $P(y_j |y_i, \mathcal{S})$ for simplicity. These probabilities will be also denoted as $V_{i,j}^{\mathcal{S}}$ when referring to them as variables in the linear program. All these values will directly define the transition matrix $T$ in the following way:
\filbreak

\begin{equation}
t_{i,j} = \sum_{\mathcal{S}\subseteq Y} P(y_j | y_i, \mathcal{S})    P(\mathcal{S}|y_i) = \sum_{\mathcal{S}\subseteq Y}  V_{i,j}^\mathcal{S} P(\mathcal{S} | y_i).   
\end{equation}

As we assume that the ground-truth class of an input is always reachable, for the inputs of class $y_i$, the probabilities corresponding to those sets $\mathcal{S}$ in which $y_i \notin \mathcal{S}$ will be zero. That is, $p(\mathcal{S}|y_i)=0$ if $y_i\notin \mathcal{S}$. Similarly, $P(y_j | y_i, \mathcal{S})$ must be zero if $y_j \notin \mathcal{S}$.

In order to find the appropriate values for the variables $V_{i,j}^{\mathcal{S}}$, we will solve the following linear program:

\begin{equation}
\label{eq:linear_program_crm}
\begin{aligned}
\textrm{min} & \quad  z = \sum_{i=1}^k t_{i,i} \\
\textrm{s.t.} &  \quad \mathcal{P}(Y)\cdot T=\widetilde{\mathcal{P}}(Y) \\
&  \quad \sum_{j=1}^k V_{i,j}^\mathcal{S} = 1 & \forall i\!\in\!\{1,\dots, k \} \ , \ \forall \mathcal{S} \! \subseteq \! Y \\
& \quad 0 \leq  V_{i,j}^\mathcal{S} \leq 1   & \forall i,j\!\in\!\{1,\dots, k \} \ , \ \forall \mathcal{S} \! \subseteq \! Y\\
& \quad V_{i,j}^\mathcal{S}=0  & y_j\notin \mathcal{S}.\\
\end{aligned}
\end{equation}

The main disadvantage of this method is that it requires a considerably larger number of decision variables, bounded by $O(2^kk^2)$, assuming that for $k$ classes there are $2^k$ possible subsets of reachable classes $\mathcal{S}$, each with an associated probability $P(\mathcal{S}|y_i)$, and for each of them another distribution of $k$ probabilities $P(y_j|y_i,\mathcal{S})$, which are optimized in the linear program. 

Due to the high number of possible subsets, in practice, $P(\mathcal{S}|y_i)$ will be zero for many of the subsets $\mathcal{S}$. This reduces the number of parameters that can be tuned, and also, as a consequence, the number of probability distributions that can be produced. For this reason, to avoid having multiple null values for those probabilities, in this method we will smooth every probability distribution $P(\mathcal{S}|y_i)$ using the Laplace smoothing \citep{manning2008introduction}.

In addition, after a preliminary experiment we discovered that, because of the values of $P(\mathcal{S}\!=\!\{y_i\} | y_i)$, the linear problem was infeasible for many target probability distributions, especially for low distortion thresholds. This is because the values $t_{i,i}$ are highly influenced by such probabilities, which, indeed, are lower thresholds for $t_{i,i}$. In addition, those probabilities can be considerably higher than those corresponding to the remaining subsets if there is a sufficiently large proportion of samples that can not be fooled, especially for low values of $\epsilon$. This also translates into a low fooling rate expectation.

To avoid all these consequences, after the Laplace smoothing, we set every ${P(\mathcal{S}=\{y_i\} | y_i)}$ to zero and normalize every distribution ${P(\mathcal{S} | y_i)}$ accordingly, $i=1,\dots,k$, even if this can reduce the effectiveness of the method in producing the target probability distribution, as we are not considering the estimated proportion of samples that can not be fooled.

\subsection{Overview of the Attack Strategies}
\label{sec:comparing_methods}
All the strategies introduced in the previous sections can be used to generate the transition matrices needed to produce adversarial class probability distributions, all of them relying on a different strategy to model the solutions.

Both the UBM and EWTM provide a simple framework that allows the transition matrices $T$ to be directly optimized. In the UBM, the proportion of samples $r_{i,j}'$ that can be moved from each class $y_i$ to another class $y_j$ is estimated, and those values are used as upper bounds for $T$, assuming that, in practice, it will be unlikely to move a larger proportion of samples. In the EWTM, the aim is to transform the transition matrix $\hat{R}$ in order to meet our particular requirements, without setting boundaries to the values of $T$. A positive point in both methods is the low number of parameters to be optimized, bounded by $O(k^2)$. The CRM, however, requires a considerably larger number of parameters, bounded by $O(2^kk^2)$, but provides a more comprehensive and general approach to generate the transition matrix. In particular, contrarily to the previous strategies, it allows the particular set of reachable classes for each instance individually to be taken into account, instead of considering \textit{aggregated} information.

\section{Validating Our Proposals: Setup and Results}
\label{sec:validation}
In this section, we present the particular task, dataset, model and further details regarding the experimental setup used to validate our proposals. We also report the obtained results, in which we measure the effectiveness of the introduced approaches according to different criteria.\footnote{Our code is available at: \url{https://github.com/vadel/ACPD} (see Appendix \ref{app:reproducibility} for further details).}

\subsection{Case of Study: Speech Command Classification}
\label{sec:problem}
Due to advances in automatic speech recognition technologies based on machine learning models, and their deployment in smartphones, voice assistants and industrial applications, there has been a rapid increase in the study of adversarial attacks and defenses for such models \citep{carlini2018audio,li2019adversarial,michelkoerich2020crossrepresentation,subramanian2020study,sallo2021adversarially,esmaeilpour2021cyclic}, despite being considerably less studied than computer vision problems. For these reasons, we have decided to validate our proposal in the task of speech command classification, an exemplary and representative task in this domain.\footnote{
Nevertheless, we remark that our methods can be directly applied to any problem or domain, as long as it is possible to generate targeted adversarial attacks, and that this selection is only for illustration purposes.
}

We use the Speech Command Dataset \citep{warden2018speech}, which consists of a set of WAV audio files of 30 different spoken commands. The duration of all the files is fixed to 1 second, and the sample-rate is 16kHz in all the samples, so that each audio waveform is composed of $16000$ values, in the range $[-2^{15},2^{15}]$. We use a subset of ten classes, following previous publications \citep{warden2018speech,alzantot2018did,yang2019characterizing,du2020sirenattack,gong2019realtime,li2020advpulse}, so that our results are more comparable with previous works in the literature:
\textit{Yes, No, Up, Down, Left, Right, On, Off, Stop}, and \textit{Go}. In order to provide a more realistic setup, two special classes have also been considered: \textit{Silence}, representing that no speech has been detected, and \textit{Unknown}, representing an unrecognized spoken command, or one which is different to those mentioned before. The dataset contains 46.258 samples, accounting for approximately 13 hours of data, and it is split into training (80\%), validation (10\%) and test (10\%) sets, following the standard partition procedure proposed in \cite{warden2018speech}.

Also following previous publications \citep{warden2018speech,alzantot2018did, du2020sirenattack,gong2019realtime,li2020advpulse}, a Convolutional Neural Network will be used as a classification model, based on the architecture proposed in \citet{Sainath2015convolutional}, which is particularly well-suited for small-footprint keyword recognition tasks. The test accuracy of the model is 85.52\%. The input of the model will be the MFCC coefficients extracted from the raw audio waveform, which is a standard feature extraction process in speech recognition \citep{muda2010voice}. Nevertheless, the adversarial examples will be generated directly in the audio waveform representation, as done in previous works \citep{carlini2018audio,qin2019imperceptible,alzantot2018did,du2020sirenattack,yakura2018robust}.

\subsubsection{Experimental Details}
\label{sec:experimental_details}
The ultimate goal is to validate that any desired probability distribution $\widetilde{\mathcal{P}}(Y)$ can be approximated with a low error by guiding a targeted adversarial attack using Algorithm \ref{algorithm_general_deepfool} and a transition matrix $T$, which has been optimized using any of the methods introduced in Section \ref{sec:methods}.

To show that any targeted attack strategy can be extended, we will evaluate our methods using a wide range of attacks, which have been exhaustively employed in the literature:  DeepFool \citep{moosavi2016deepfool}, Fast Gradient Sign Method \citep{goodfellow2014explaining}, Projected Gradient Descent \citep{madry2018deep} and Carlini \& Wagner attack \citep{carlini2017towards}. A brief introduction to these algorithms is provided in Appendix \ref{sec:supp_description_attacks}. For the sake of simplicity, the experimental results presented in this section will be reported for the DeepFool algorithm\footnote{In order to fit in our specification, we employed a targeted version of DeepFool, as described in Appendix \ref{sec:supp_description_deepfool}.}, whereas the results obtained with the other attack algorithms will be reported in Appendix \ref{sec:supp_results_other_attacks}.

In all the experiments, we will assume a uniform initial probability distribution $\mathcal{P}(Y)$. In Section \ref{sec:results_particular_case}, the particular case in which $\widetilde{\mathcal{P}}(Y)=\mathcal{P}(Y)$ will be tested, that is, when the aim is to reproduce the original probability distribution $\mathcal{P}(Y)$ obtained by the model (in our case the uniform distribution). Afterward, in Section \ref{sec:results_general_case}, a more general scenario will be tested, in which different target probability distributions $\widetilde{\mathcal{P}}(Y)$ will be randomly sampled from a Dirichlet distribution of $k=12$ parameters and $\alpha_i=1 , \  1 \leq i \leq 12$. A total of 100 different target probability distributions will be sampled, and our methods will be tested in each of them. This general case will be used to provide an exhaustive comparison of the effectiveness of the methods introduced.

To generate the transition matrices $T$, a set of samples $\mathcal{X}$ will be used, composed of 500 samples per class, which makes a total of 6000 input samples. In particular, $\mathcal{X}$ will be used to generate the auxiliary matrix $R$ required in the UBM and EWTM, as described in Equation \eqref{eq:matrix_r}, and, for the case of the CRM, to estimate the probabilities described in Equation \eqref{eq:m4_Psi}. The generated transition matrices $T$ will be tested using another set of samples $\hat{\mathcal{X}}$, disjoint from $\mathcal{X}$, also composed of 500 inputs per class. The proportion of samples that has been classified as each particular class after the attack is applied to every input in $\hat{\mathcal{X}}$ will be taken as the \textit{empirical} probability distribution, and will be denoted $\hat{\mathcal{P}}(Y)=( \hat{p}_1, \dots, \hat{p}_k )$. Using this collection of data, we will evaluate to what extent the empirical probability distributions $\hat{\mathcal{P}}(Y)$ match $\widetilde{\mathcal{P}}(Y)$. The similarity between both distributions will be measured using different metrics: the maximum and mean absolute difference, the Kullback-Leibler divergence and the Spearman correlation.

To thoroughly evaluate our methods, we randomly sampled a set $\bar{\mathcal{X}}$ of 1000 inputs per class from the training set of the Speech Command Dataset, and computed a 2-fold cross-validation, using one half of $\bar{\mathcal{X}}$ as $\mathcal{X}$ and the other half as $\hat{\mathcal{X}}$. Moreover, we launched 50 repetitions of the cross-validation process, using in every repetition a different random partition of $\bar{\mathcal{X}}$. An additional evaluation of our methods considering different sizes for the set $\mathcal{X}$ will be provided in Appendix \ref{sec:supp_less_train}.

The transition matrices will be generated using the four linear programs described in Sections \ref{sec:method1}, \ref{sec:method2}, \ref{sec:method3} and \ref{sec:method4}.
The linear programs are solved using the Python PuLP library \footnote{\url{https://github.com/coin-or/pulp}} and the Coin-or Branch and Cut (CBC) solver \footnote{\url{http://www.coin-or.org/}}.
For the AM and the UBM, an upper bound of $\xi=0.01$ will be set for the values in $L$. For the AM, the UBM and the EWTM, $\gamma_1=\gamma_2=1$ will be set, as well as $\gamma_3=10$ for the UBM, in order to avoid the relaxation of the upper-bounds $r'_{i,j}$.

In addition, the $\ell_2$ norm of the adversarial perturbation (in the raw audio waveform representation) will be used as the distortion metric $\varphi$. The results will be computed under the following maximum distortion thresholds: $\epsilon\in \{0.0005,$ $0.001,$ $0.0025,$ $0.005,$ $0.01,$ $0.05,$ $0.1,$ $0.15 \}$.
These values were empirically selected in order to evaluate the behavior of our methods depending on how restricted the adversary is, ranging from scenarios where only very few class transitions can be performed (i.e., low values of $\epsilon$), to scenarios where the number of possible transitions is high (i.e., high values of $\epsilon$).

Finally, our methods will be compared against two baseline methods. With the first baseline, for each input $x$, the target class will be selected according to the probabilities defined in the target distribution $\widetilde{\mathcal{P}}(Y)$. Notice that, following the introduced methodology, this method can be modeled as a transition matrix $T$ in which all the rows contain the target distribution $\widetilde{P}(Y)$. Thus, this method will presumably provide a good approximation of the target distribution, but also fooling rates far from the optimum, as it only focuses on producing the target distribution, with no particular incentive to maximize the fooling rate. Hence, we will refer to this baseline as the \textit{Maximum Approximation Baseline} (MAB). 

On the other hand, the second baseline will also follow the same strategy as the MAB, with the difference that, in order to maximize the fooling rate of the attack, the diagonal of $T$ (i.e., the probability of staying in the ground-truth class) will be set to zero, and each row will be normalized accordingly:
\begin{equation}
\begin{cases} 
t_{i,i} = 0, & 1\leq i \leq k,\\
t_{i,j} = \dfrac{\widetilde{p}_j}{\sum_{\substack{\\r=1\\r\neq i}}^k \widetilde{p}_r} ,  &  1\leq i,j \leq k, \ \ i\neq j.
\end{cases} 
\end{equation}
Therefore, this baseline provides a maximum fooling rate, but, presumably, at the expense of producing worse approximations to $\widetilde{\mathcal{P}}(Y)$ than the MAB. For this reason, we will refer to this baseline as the \textit{Maximum Fooling Rate Baseline} (MFRB).

For the purpose of evaluating the baselines under the same conditions as our methods, the normalization described in Equation \eqref{eq:normalization} will be also employed before each attack (see Algorithm \ref{algorithm_general_deepfool}, lines 11-21). In this way, sampling a target class that is not reachable from the input $x$ at hand is avoided, which favors the baselines.

\subsection{Illustrative Case: Reproducing the Initial Probability Distribution}
\label{sec:results_particular_case}
For illustration purposes, we first report the results obtained for the particular scenario in which we want to produce the same probability distribution that the model produces when it is applied on clean samples. 
Notice that this distribution is the same as the ground truth distribution of the classes, since we assume that the model produces a correct classification for the original samples. Having the ability to reproduce such distributions allows an adversary to deploy attacks that are less likely to be detected in the long run, for instance, by label-shift detection methods that can warn against the presence of multiple adversarial attacks against the model \citep{rabanser2019failing}.\footnote{We clarify that this does not imply that each individual attack that is sent to the model will also be less detectable, as this depends on the underlying targeted adversarial attack method employed.}

\begin{figure}
\centering
\includegraphics[scale=0.67]{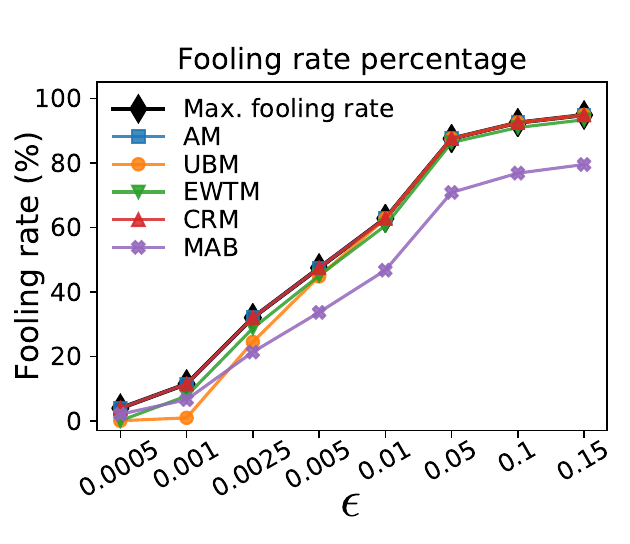}
\ \ \ \ \includegraphics[scale=0.67]{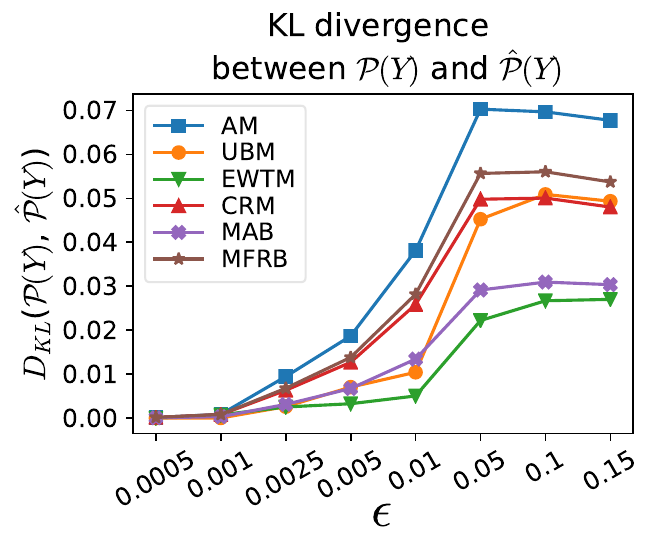} 
\caption{Fooling rates (left) and Kullback-Leibler divergence (right) obtained with each of the proposed methods in 50 2-fold cross-validation trials, for the particular case in which the target distribution $\widetilde{\mathcal{P}}(Y)$ is the initial  (uniform) probability distribution $\mathcal{P}(Y)$. The results corresponding to the MFRB have been omitted from the left figure, since that method achieves the maximum fooling rate by definition.}
\label{fig:targeting_initial_metrics}
\end{figure}

To begin with, Figure \ref{fig:targeting_initial_metrics} (left) shows the achieved fooling rates for each method, as well as the maximum fooling rate that can be achieved for every $\epsilon$ as reference, that is, the percentage of inputs in $\hat{\mathcal{X}}$ for which it is possible to create a targeted attack capable of fooling the model. These results have been averaged for the 50 different 2-fold cross-validations. For the sake of simplicity, and since the MFRB achieves by definition the maximum possible fooling rate, the results corresponding to that baseline are not included in the figure. According to the results, the four attack methods maintained fooling rates very close to the optimal values independently of the distortion threshold, with the exception of the UBM and the EWTM, in which a loss can be observed (of approximately 10\% and 4\%, respectively) for the lowest values of $\epsilon$ evaluated. It can also be noticed that the lowest fooling rates are achieved by the MAB, with a loss of approximately 15\% for $\epsilon \geq 0.005$.

\begin{figure}
\centering
\includegraphics[scale=0.385]{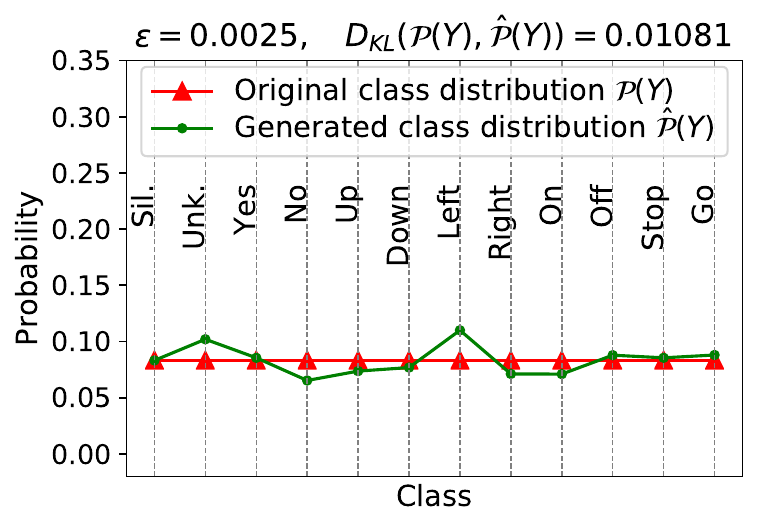}
\includegraphics[scale=0.385]{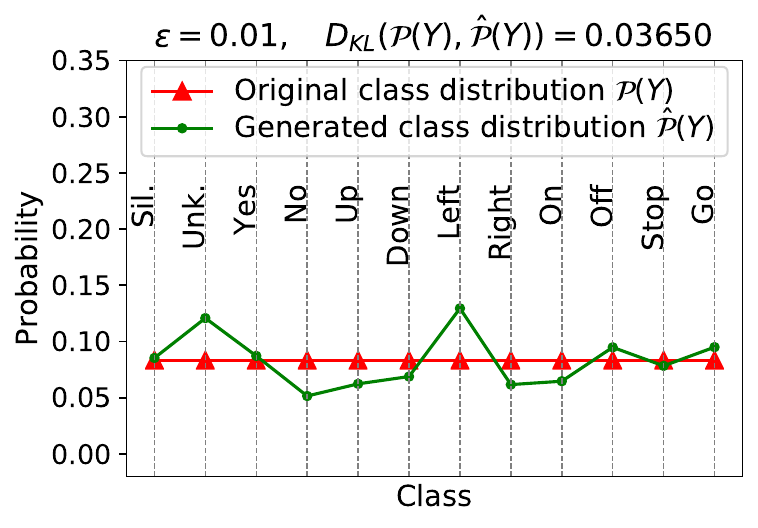}
\includegraphics[scale=0.385]{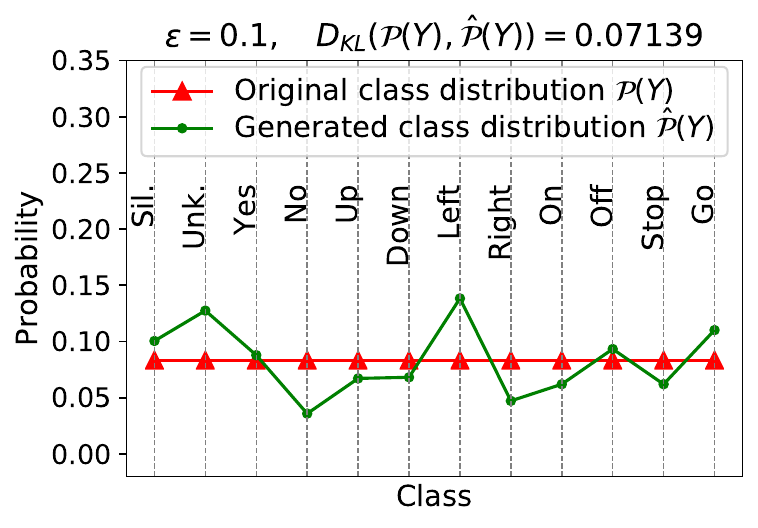}
\includegraphics[scale=0.385]{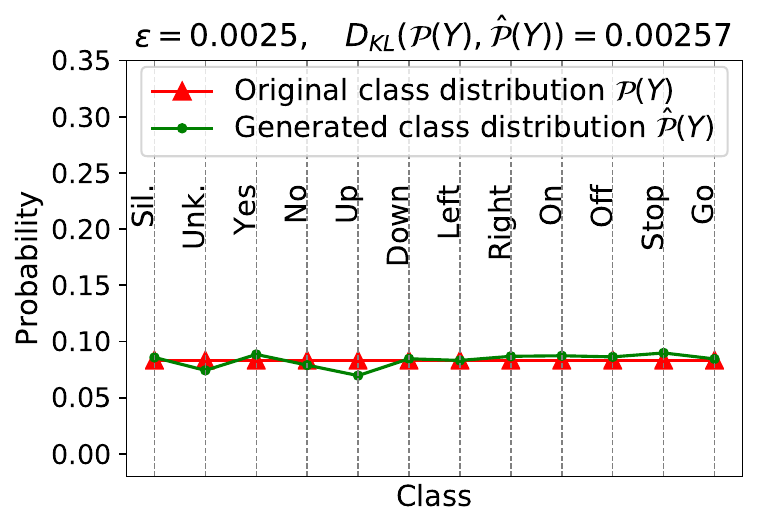}
\includegraphics[scale=0.385]{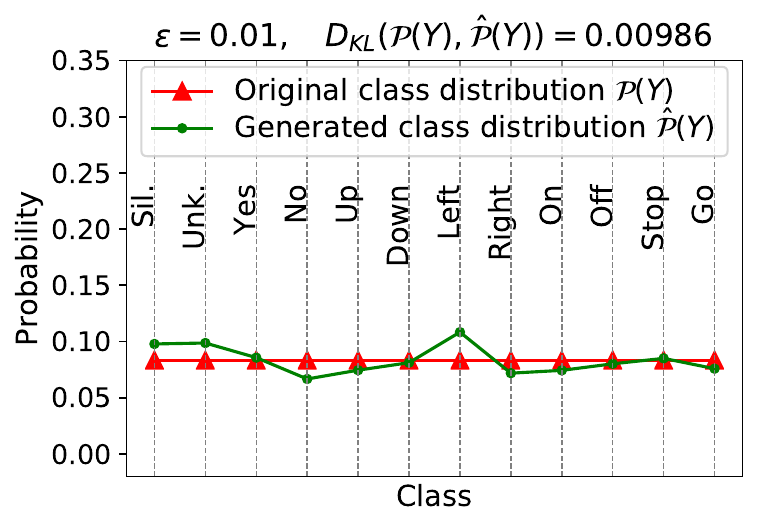}
\includegraphics[scale=0.385]{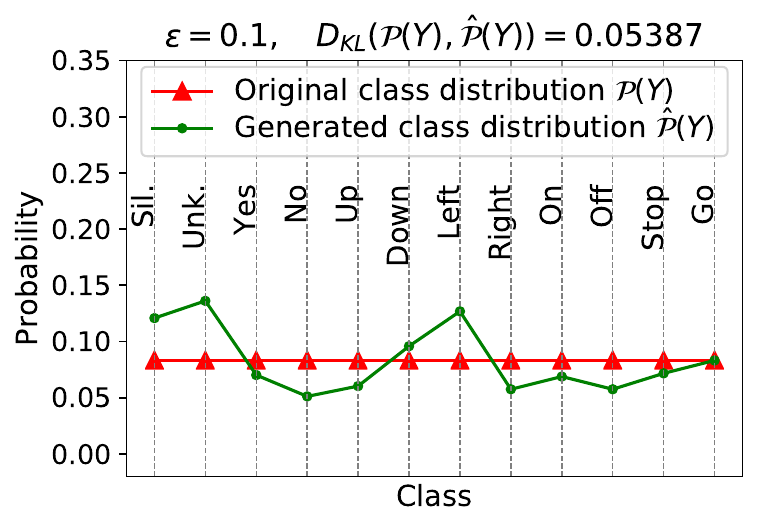}
\includegraphics[scale=0.385]{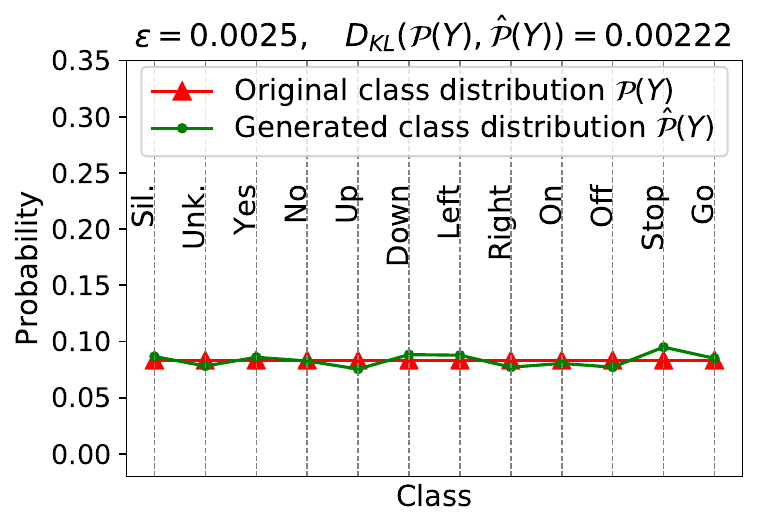}
\includegraphics[scale=0.385]{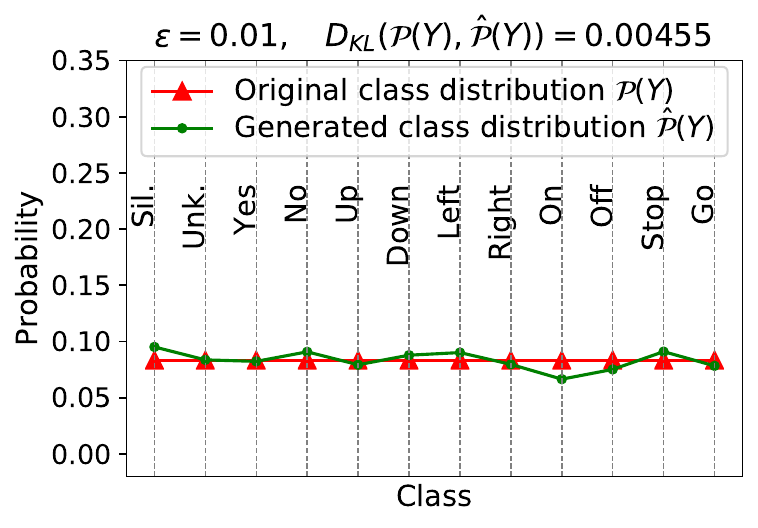}
\includegraphics[scale=0.385]{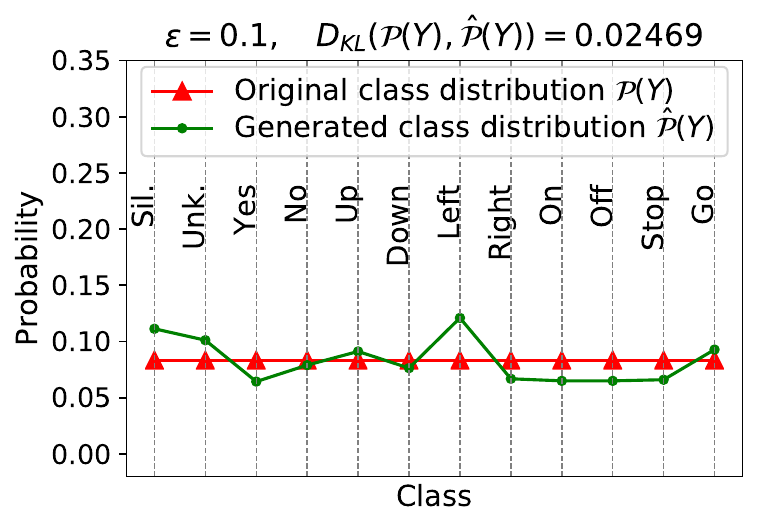}
\includegraphics[scale=0.385]{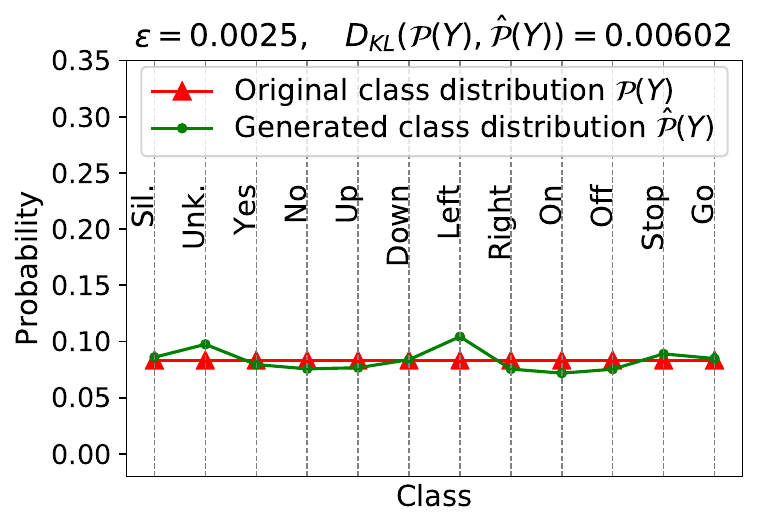}
\includegraphics[scale=0.385]{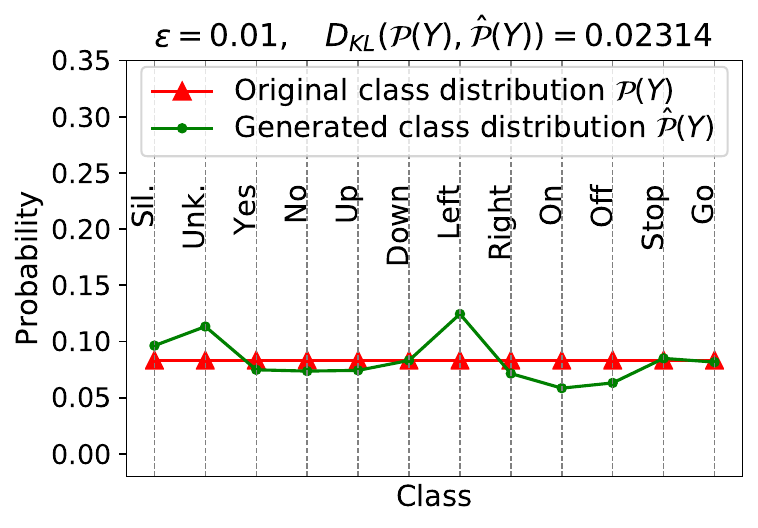}
\includegraphics[scale=0.385]{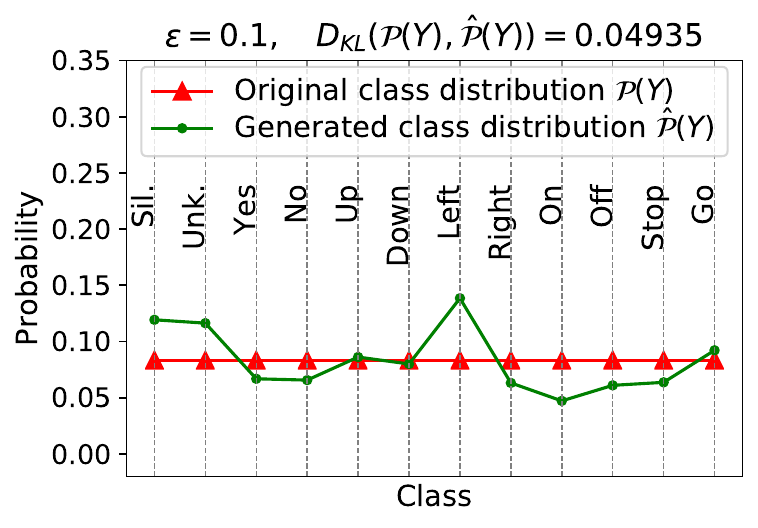}
\caption{Comparison between the target distribution (in this case the initial probability distribution) $\mathcal{P}(Y)$ and the produced probability distribution $\hat{\mathcal{P}}(Y)$, for the four different methods introduced: AM (first row), UBM (second row), EWTM (third row) and CRM (fourth row). The results are shown for three different values of $\epsilon$, and have been computed for one of the folds of the cross-validation trials. The Kullback-Leibler divergence between both distributions, $D_{KL}(\mathcal{P}(Y), \hat{\mathcal{P}}(Y))$, is also shown above each figure.}
\label{fig:targeting_initial}
\end{figure}

Regarding the effectiveness in reproducing the initial probability distribution, Figure \ref{fig:targeting_initial_metrics} (right) includes the average Kullback-Leibler divergence obtained for every $\epsilon$.
In order to better assess the similarity between the initial probability distribution and the ones produced after perturbing the inputs with our attacks, Figure \ref{fig:targeting_initial} contains a graphical comparison of these distributions, for one of the folds included in the cross-validation trials, considering three different maximum distortion thresholds $\epsilon$. These figures also include the Kullback-Leibler divergences between both distributions, as a reference to compare the value of this metric and the similarity between the perturbations. According to the results, in all the cases the algorithms were able to maintain a probability distribution very close to the original one,  the EWTM being the most accurate, the AM the least accurate, and the remaining approaches achieving intermediate results.

It is noteworthy that, in this particular case, the obtained approximations of the target probability distributions are more accurate for the lowest $\epsilon$ values tried. This is due to the fact that, for low distortion thresholds, the number of inputs for which the model can be fooled is lower, and therefore, a larger number of inputs remains correctly classified as their ground-truth class, which makes the empirical probability distribution $\hat{\mathcal{P}}(Y)$ closer to the original.  However, note that the results obtained for high values of $\epsilon$ also represent close approximations of the target distributions, and at the same time, the model is fooled for almost all the input samples.

\newpage

\subsection{Deeper Exploration}
\label{sec:results_general_case}
In this section, we provide a deeper evaluation of our methods, testing them against 100 probability distributions, randomly drawn from a Dirichlet distribution, as described in Section \ref{sec:experimental_details}.

First, we compute the percentage of cases in which the methods managed to generate a valid transition matrix, that is, one which satisfies all the restrictions of the corresponding linear program. This information is shown in Table \ref{tab:success_perc}, for different values of $\epsilon$. Note that the baselines were not considered for this analysis, since they do not require solving a linear program. In particular, for each method, the values in Table \ref{tab:success_perc} represent the percentage of cross-validation trials in which a valid transition matrix was found, averaged for the 100 target distributions. If a method failed in any of the folds of a cross-validation trial, a failure is reported for the whole cross-validation trial. According to the results, the AM, the UBM and the CRM managed to create a valid matrix for all the cases tried, independently of the distortion threshold. For the EWTM, although it also achieved a total success for values of distortion above or equal to 0.0025, the percentage drops to 38.8\% for $\epsilon=0.0005$. This is due to the larger number of zeros in the matrices $\hat{R}$ for such low distortion thresholds, which makes it impossible to find feasible solutions through an element-wise multiplication with another matrix.

Table \ref{tab:success_perc} also includes the success percentages of one variant of  the UBM and two variants of the CRM. Regarding the UBM, without relaxing the upper bounds of the transition matrix $T$, 100\% success is achieved for values of distortion $\epsilon>0.05$, but the percentage of cases for which a valid transition matrix was found drops dramatically for lower values of $\epsilon$.  Regarding the CRM, without the Laplace smoothing and without fixing the probabilities $P(\{y_i\}|y_i)$ to zero, the method was not able to generate a valid transition matrix for distortions below $0.05$, and even in the maximum distortion threshold tried the method only succeeded in 46.2\% of the cases. Applying the Laplace smoothing (without fixing $P(\{y_i\}|y_i)=0$), those results improve significantly, particularly for the highest distortion thresholds tried, succeeding in more than approximately 80\% of the cases for $\epsilon\geq$ 0.1, and in 70.6\% of the cases for $\epsilon=0.05$. These results clearly reflect that those corrections are necessary to make the linear programs feasible.

\begin{table}[]
\centering
\begin{threeparttable}
\begin{tabular}{@{}lcccccccc}
\toprule
     & \multicolumn{8}{c}{Maximum distortion amount ($\epsilon$)}  \\
\cmidrule{2-9}
     Method   & {$0.0005$} & {$0.001$} & {$0.0025$} &{$0.005$} & {$\ 0.01 \ $} & {$ \ 0.05 \ $} & {$\ \ 0.1 \ \ $} & {$\ \ 0.15 \ $} \\ \midrule
AM & {100.0} & {100.0} & {100.0} & {100.0}  & {100.0}  & {100.0} & {100.0}  & {100.0} \\
UBM & {100.0} & {100.0} & {100.0} & {100.0}  & {100.0}  & {100.0} & {100.0}  & {100.0} \\
EWTM & \: 38.8 & \: 99.9 & {100.0} & {100.0}  & {100.0}  & {100.0} & {100.0}  & {100.0} \\
CRM & {100.0} & {100.0} & {100.0} & {100.0}  & {100.0}  & {100.0} & {100.0}  & {100.0} \\
\midrule
UBM \tnote{1} & \ \; 0.0 & \ \; 0.0 & \ \; 0.0 & \ \; 9.2  & \: 67.9  & \: 99.9 & 100.0  & 100.0 \\
CRM \tnote{2} & \ \; 0.0 & \ \; 0.0 & \ \; 0.0 & \ \; 0.0 & \ \; 0.0 & \: 14.0 & \: 30.3 & \: 46.2  \\
CRM \tnote{3} & \: 10.6 & \: 14.0 & \: 20.0 & \: 25.7 & \: 37.5 & \: 70.6 & \: 79.7 & \: 85.3  \\
 \bottomrule
\end{tabular}
\begin{tablenotes}
\item [1] Without relaxing the upper bound restrictions of $T$ using the auxiliary decision variable $\eta$.
\item [2] Without the Laplace smoothing and without fixing the values of $P(\{y_i\}|y_i)$ to zero.
\item [3] Without fixing the values of $P(\{y_i\}|y_i)$ to zero.
\end{tablenotes}
\end{threeparttable}
\caption{Success percentages in generating valid transition matrices for the different methods introduced.}
\label{tab:success_perc}
\end{table}

Secondly, the average fooling rates obtained by each method is compared in Table \ref{tab:fooling_rates}, for each value of $\epsilon$. In addition, the table includes, for reference purposes, the maximum fooling rate that can be obtained with a maximum distortion $\epsilon$.
All the values have been averaged for the 100 target probability distributions considered in the experiment and for the 50 2-fold cross-validations carried out for each of them.\footnote{The cross-validation processes in which a method failed in generating a valid matrix $T$ for any of the two folds were discarded, and, therefore, the results might be slightly biased for the EWTM and $\epsilon < 0.001$.} 
The results demonstrate that, whereas, by construction, the MFRB always achieves the optimum fooling rate, the MAB achieves the worst results in the majority of the cases, of approximately 15\% below the optimum for $\epsilon\geq 0.005$. In contrast, a very high fooling rate is maintained in the AM, the UBM (for $\epsilon > 0.01$) and the CRM, with a negligible loss with respect to the maximum achievable value. The EWTM, however, achieved slightly lower fooling rates, of approximately 8\% below the maximum, independently of the distortion threshold. A similar loss is observed for the UBM when $\epsilon\leq 0.01$.

\begin{table}[]
\centering
\begin{tabular}{@{}lS[table-format=1.2]S[table-format=2.2]S[table-format=2.2]S[table-format=2.2]S[table-format=2.2]S[table-format=2.2]S[table-format=2.2]S[table-format=2.2]}
\toprule
& \multicolumn{8}{c}{Maximum distortion amount ($\epsilon$)}  \\
\cmidrule{2-9}
        & {$0.0005$} & {$\ 0.001$} & {$0.0025$} &{$0.005$} & {$\ 0.01 \ $} & {$\ 0.05 \ $} & {$\ \  0.1 \ \ $} & {$\ \  0.15 \ \ $} \\ \midrule
AM   & 3.80 & 11.17 & 31.58 & 46.98 & 62.36 & 87.29 & 92.31 & 94.69\\
UBM  & 0.45 &  2.88 & 19.06 & 38.03 & 57.89 & 87.05 & 92.28 & 94.68 \\
EWTM & 1.88 &  6.87 & 23.59 & 38.65 & 53.60 & 79.66 & 85.21 & 87.84\\
CRM  & 3.90 & 11.29 & 31.55 & 46.88 & 62.23 & 87.26 & 92.31 & 94.70\\
\midrule
MAB  & 2.06 &  6.55 & 21.33 & 33.72 & 46.87 & 71.02 & 76.96 & 79.64 \\
MFRB & 3.93 & 11.47 & 32.02 & 47.44 & 62.80 & 87.54 & 92.48 & 94.86  \\
\midrule
Max. FR & 3.93 & 11.47 & 32.02 &  47.44 &  62.80 & 87.54 & 92.48 &  94.86\\
 \bottomrule
\end{tabular}
\caption{Fooling rate (FR) percentages achieved by the different methods introduced.}
\label{tab:fooling_rates}
\end{table}

\begin{figure}
\centering
\includegraphics[scale=0.63]{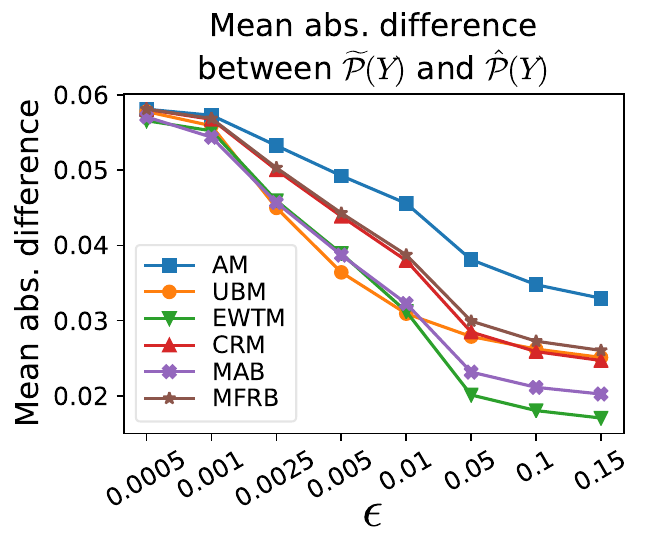} \hspace{0.2cm}
\includegraphics[scale=0.63]{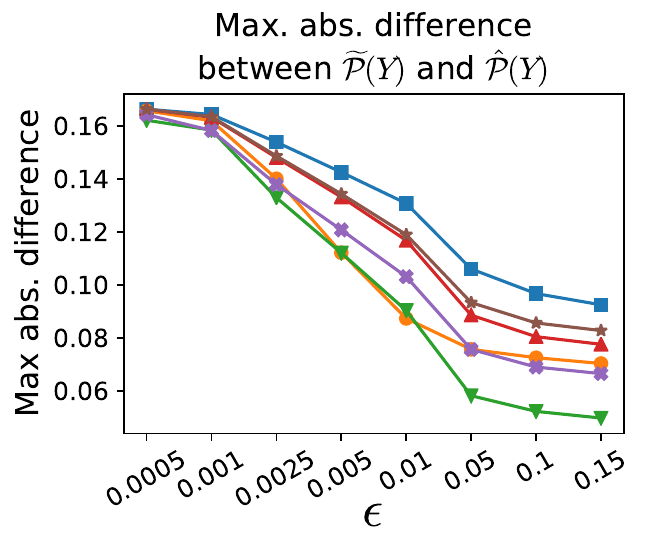}  
\includegraphics[scale=0.63]{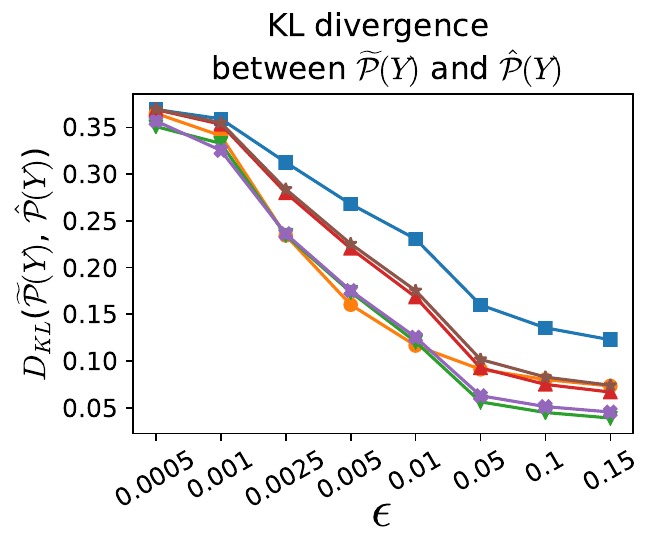} \hspace{0.2cm}
\includegraphics[scale=0.63]{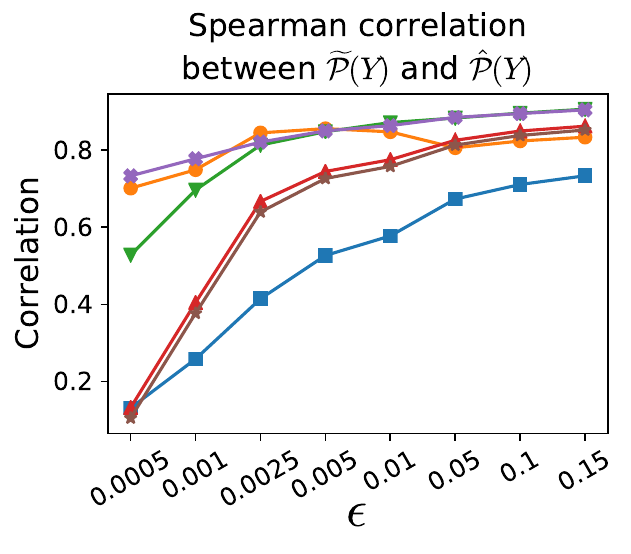} 
\caption{Sensitivity analysis of different similarity metrics between the produced probability distribution $\hat{\mathcal{P}}(Y)$ and the target probability distribution $\widetilde{\mathcal{P}}(Y)$: mean absolute difference, maximum absolute difference, Kullback-Leibler (KL) divergence and Spearman correlation.}
\label{fig:full_metric_comp}
\end{figure}

To conclude the analysis, the average similarity between the target distributions $\widetilde{\mathcal{P}}(Y)$ and the corresponding empirical distributions $\hat{\mathcal{P}}(Y)$ is analyzed in Figure \ref{fig:full_metric_comp}, independently for the different similarity metrics considered and for every maximum distortion threshold. 
First of all, it is clear that the effectiveness of the methods in reproducing the target distribution increases with the maximum allowed distortion. Apart from that, it can be seen that the AM achieves worse results compared to the rest, thereby validating the hypothesis that the more informed strategies employed in the UBM, EWTM and CRM are capable of increasing the effectiveness of the attack. Indeed, analyzing the results obtained with the remaining methods, the maximum absolute difference between the probabilities of the classes is below $0.09$ for $\epsilon \geq 0.05$, which reflects a very high similarity. In fact, for the mean absolute difference, this value decreases to $0.03$. The Kullback-Leibler divergence also shows the same descending trend as the maximum and mean differences. Finally, the Spearman correlation between both distributions is above $0.80$ for $\epsilon\geq 0.05$, which indicates that even if there are differences between the values, both distributions are highly correlated. 

Comparing the overall effectiveness of the methods in approximating the target distributions, the UBM and the EWTM were the most effective for low and intermediate distortion thresholds ($\epsilon \leq 0.01$), followed by the MAB, while the CRM and the MFRB achieved intermediate results. For high distortion thresholds ($\epsilon \geq 0.05$), in contrast, the EWTM achieved the best results with a notable margin with respect to the other methods, which show a more similar performance.

Comparing our methods with the baselines, on the one hand, the MAB achieves results competitive with the UBM and the EWTM in terms of approximating the target distribution.
Nevertheless, it can be noticed that the MAB is outperformed by the EWTM in most cases, and even by the UBM for intermediate values of $\epsilon$, while it is also outperformed in terms of fooling rate by all the remaining methods, with a considerable margin (as shown previously in Table \ref{tab:fooling_rates}). Hence, the MAB is dominated by our methods in both factors. On the other hand, whereas the MFRB cannot be outperformed in terms of fooling rate (since it guarantees the optimal value), it is outperformed in terms of the quality of the approximation by our methods.
These results corroborate that the proposed methods are capable of taking advantage of the information about the problem provided in order improve their joint effectiveness in the two main goals of the attack: closely approximating the target distribution for the classes while keeping remarkable effectiveness in the objective of fooling the model for any incoming input sample.

Finally, as an overview of the distortion, Table \ref{tab:db_metrics} shows the average distortion level introduced by the perturbations, in decibels (dB). Following the methodology introduced in previous related works on adversarial perturbations in speech signals \citep{carlini2018audio, neekhara2019universal,xie2020realtime,vadillo2019universal}, the distortion has been computed as 
\begin{equation}
\label{eq:db_metric}
{dB(x,v)=\max_i 20\cdot\log_{10}\left(|v_i|\right) - \max_i 20\cdot\log_{10}\left(|x_i|\right)},
\end{equation}
$x$ being the clean signal and $v$ the perturbation.\footnote{Notice that the metric described in Equation \eqref{eq:db_metric} is used for a post-hoc analysis and not to optimize the adversarial attacks, for which the $\ell_2$ norm was used, as described in Section \ref{sec:experimental_details}.}
According to this metric, the lower the value, the less perceptible the perturbation. 
Even for the highest values of $\epsilon$ tried, the mean distortion level is far below -32dB, which is the maximum acceptable distortion threshold assumed in related works \citep{carlini2018audio, neekhara2019universal, vadillo2019universal}. To empirically assess the imperceptibility of the adversarial perturbations, a randomly sampled collection of our adversarial examples can be found in our webpage \footnote{\url{https://vadel.github.io/acpd/AudioSamples.html}}.

\begin{table}[]
\centering
\scalebox{0.9}{
\begin{tabular}{@{}ccccccccc@{}}
\toprule
$\epsilon$ & $0.0005$  & $0.001$  & $0.0025$ & $0.005$ & $0.01$ & $0.05$ & $0.1$ & $0.15$ \\ \midrule
dB & -80.69 & -78.21  & -72.73 & -69.13 & -65.60 & -58.30 & -55.82 & -54.61 \\
\bottomrule
\end{tabular}
}
\caption{Average distortion levels introduced by the adversarial perturbations generated, measured in decibels (dB).}
\label{tab:db_metrics}
\end{table}

\subsection{General Comparison of the Introduced Approaches}

As a general overview of the effectiveness of the introduced strategies, focusing on the UBM, the EWTM and the CRM, the three of them provided an effective way to find optimal transition matrices, capable of producing the desired target probability distributions. In addition, and considering that the effectiveness of the methods depends on multiple factors, there is no one best method in all the cases. For instance, the EWTM was overall the most effective one in approximating the desired probability distributions, but achieved lower fooling rates than the UBM and the CRM, which achieved values close to the maximum fooling rates. 

\begin{figure}[]
\centering
\ \ \ \ \ \, \, \includegraphics[scale=0.72]{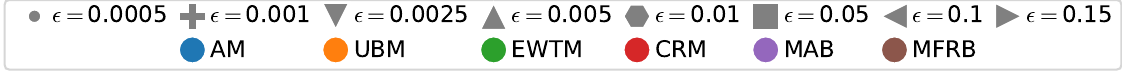}
\includegraphics[scale=0.6]{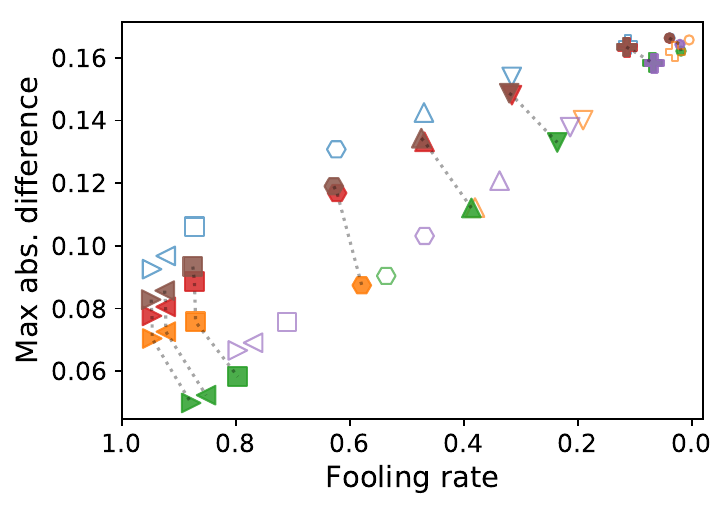} \ \ \
\includegraphics[scale=0.6]{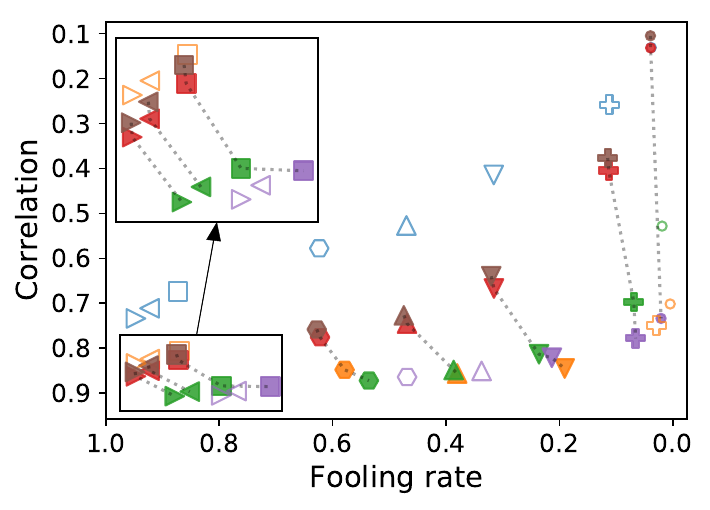} 
\includegraphics[scale=0.6]{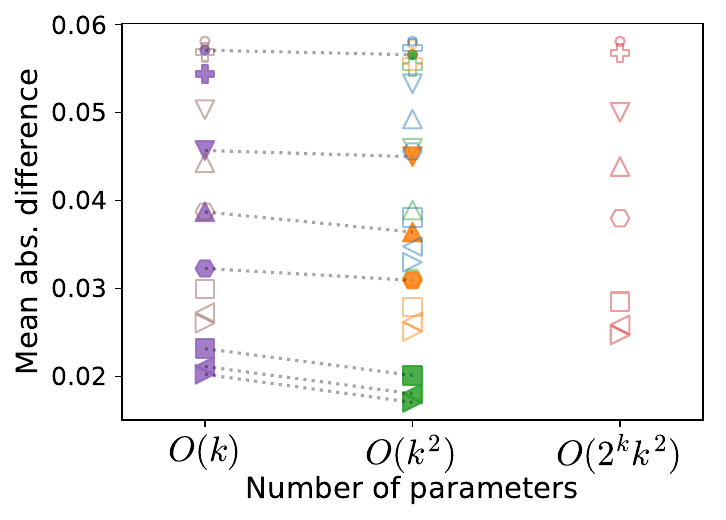}
\caption{Multifactorial comparison of the effectiveness of the six methods evaluated. For a clearer visualization, dominated values (i.e., those corresponding to methods which are outperformed in all factors by at least another method, under the same distortion threshold) have been displayed in white, whereas the non-dominated values corresponding to the same distortion threshold have been connected by dashed gray lines.}
\label{fig:pareto}
\end{figure}

This can be assessed more clearly in Figure \ref{fig:pareto}, in which a graphic comparison of the effectiveness according to the most relevant factors is provided. For a clearer visualization, dominated values (i.e., those corresponding to methods which are outperformed in all factors by at least another method, under the same distortion threshold) have been displayed in white. Moreover, the non-dominated values corresponding to the same distortion threshold have been connected by dashed gray lines. Notice also that some axes are flipped to represent in all the cases that a value is better if it is closer to the bottom-left corner. As can be seen, no method is dominated by the others in all the factors or metrics considered, with the exception of the AM (which is dominated in all the cases) and the MAB (which is dominated in most of the comparisons in which the fooling rate and the similarity metrics are traded-off). Thus, the variety of methods proposed allows us to select the one that best suits the requirements of the adversary, depending on which factors are the most relevant or which are to be optimized the most.

\pagebreak

\section{Counteracting Label-shift Detection Algorithms in Data Streaming Scenarios}
\label{sec:exp_shift_detection}
As discussed in Section \ref{sec:introduction}, a change in the probability distribution of the classes can lead to a change in the predictive performance of the models or to ethical issues \citep{vucetic2001classification,saerens2002adjusting,lipton2018detecting,biswas2021ensuring}. Therefore, some approaches have been proposed to detect and correct those shifts. In this section, we show the effectiveness of our method in producing a label-shift, even when a label-shift detection method is enforced.

The assumed scenario is as follows. First, we consider a classification model in its deployment phase, which receives a set of unlabeled instances every time unit. We also assume that the initial probability distribution of the classes $\mathcal{P}(Y)$ is known. 
Finally, we consider the presence of a label-shift detector, which evaluates whether the probability distribution of the classes at the prediction phase, $\mathcal{Q}(Y)=(q_1,\cdots,q_k)$, is different from $\mathcal{P}(Y)$. We assume that this evaluation is done periodically, for instance, after receiving a certain number of new instances.
In such a scenario, the goal of our attack will be to maliciously produce a target probability distribution $\widetilde{\mathcal{P}}(Y)$, different from $\mathcal{P}(Y)$, yet preventing the change from being detected by the label-shift detection mechanism. Otherwise, the detection mechanism can alert the user about possible attacks \citep{rabanser2019failing} or trigger actions such as retraining or replacing the model. Such actions may force the adversary to recalculate the attack strategy, interrupt the attack process or cause it to fail.

For illustration purposes, we will employ the Black Box Shift Detection (BBSD) approach proposed in \citet{lipton2018detecting} as the label-shift detector. This method assumes a realistic scenario in which the probability distribution at prediction time $\mathcal{Q}(Y)$ is unknown, since only unlabeled data is observed, which is a common scenario in practice. To address such scenarios, \citet{lipton2018detecting} propose a methods-of-moments approach to consistently estimate $\mathcal{Q}(Y)$ at prediction time, based on the predictions of the classification model.\footnote{We refer the reader to the work of \citet{lipton2018detecting} for further details.}
Once the probability distribution at prediction time is estimated, the shift detection is formulated as a statistical test under the null hypothesis $\mathbf{H_0}: \mathcal{Q}(Y)=\mathcal{P}(Y)$ and the alternative hypothesis $\mathbf{H_1}: \mathcal{Q}(Y)\neq \mathcal{P}(Y)$. 
As in \cite{rabanser2019failing}, a Pearson's Chi-Squared test will be used as the statistical test to quantify the significance of the label-shift. We will consider that the null-hypothesis is rejected (i.e., the BBSD method detects a significant shift) when the p-value is below $10^{-5}$.

As the underlying task for our experiments, we will consider a Tweet emotion classification problem, which is a popular benchmark in text classification \citep{alshahrani2021optimism,wasserblat2020exploring}, streaming classification \citep{hasan2019automatic} and quantification learning scenarios \citep{gao2016classification,perez-gallego2017using}, where the probability distribution of the output classes (which might represent, for instance, the overall opinion of the population with respect to a given topic) is of paramount relevance \citep{giachanou2016it}. We selected the \textit{Emotion} dataset proposed in \citet{saravia2018carer}, which contains Tweets categorized in 6 emotions: \textit{sadness}, \textit{joy}, \textit{love}, \textit{anger}, \textit{fear} and \textit{surprise}. 
We also selected a pretrained classifier based on the popular \textit{BERT} language model \citep{devlin2019bert}, fine-tuned for this dataset.\footnote{The model is publicly available at: \url{https://huggingface.co/bhadresh-savani/bert-base-uncased-emotion}.} The resulting model achieves a 92.65\% of accuracy in the test set of the Emotion dataset.

As the underlying adversarial attack, we selected the method proposed by \cite{alzantot2018generating}.
Finally, the Levenshtein Edit Distance \citep{levenshtein1966binary}  between the original and the adversarial text was selected as the distortion metric, normalized by the length of the longest text.\footnote{A randomly sampled collection of our adversarial examples can be found in our webpage:\\ \url{https://vadel.github.io/acpd/TextSamples.html}.} We set a maximum distortion threshold of $\epsilon=0.25$.

Since we assume a label-shift detection mechanism, it is important to note that only those target distributions that are not statistically different from $\mathcal{P}(Y)$ (according to the detection method) can be targeted, to prevent the change from being detected. Thus, if a target distribution $\bar{\mathcal{P}}(Y)$ is significantly different from $\mathcal{P}(Y)$, our best option is to find another distribution which, despite being as close to $\bar{P}(Y)$ as possible, will not cause the statistical test to reject the null hypothesis. For instance, such a trade-off can be straightforwardly managed by computing the intermediate distribution
\begin{equation}
\label{eq:py_tau_tradeoff}
\widetilde{\mathcal{P}}(Y) = (1-\tau)\mathcal{P}(Y)  + \tau \bar{\mathcal{P}}(Y), \ \tau \in [0,1],
\end{equation}
and finding the maximum value of $\tau$ so that $\widetilde{\mathcal{P}}(Y)$ is not significantly different from $\mathcal{P}(Y)$.\footnote{In our experiments, the maximum value of $\tau$ was found by means of a binary search on the range $[0,1]$.}

To evaluate the effectiveness of our method, we considered three different configurations for the source probability distribution $\mathcal{P}(Y)$. First, a roughly uniform distribution will be tested, similarly to the evaluation in the previous section. Secondly, following a similar approach to \cite{lipton2018detecting}, we considered two distributions in which a probability $p_i$ is assigned to the $i$-th class and the remaining probability mass is distributed uniformly among the remaining classes. For our experiments, we will set $p_i=0.25$ and $i=\{2,4\}$, and, following the notation of \cite{lipton2018detecting}, we will refer to these distributions as Tweak-2 and Tweak-4. 

For each $\mathcal{P}(Y)$, 1000 random Dirichlet distributions were sampled as the target distributions. We ensured, using the approach described in Equation \eqref{eq:py_tau_tradeoff}, that all of the target distributions are not being identified by the label-shift detector as significantly different from the corresponding source distribution $\mathcal{P}(Y)$.\footnote{We considered a tolerance of $10^{-4}+10^{-5}$ during the sampling process.}
In addition, to generate the transition matrices, we sampled 1000 \textit{training} inputs from the dataset, with a class proportion following $\mathcal{P}(Y)$. The EWTM will be used to optimize the transition matrices in all the cases. Once the transition matrix is generated, its effectiveness will be evaluated on a different set $\hat{\mathcal{X}}$, also composed of 1000 inputs. 
For the sake of a realistic (and challenging) evaluation, the BBSD will be evaluated in cumulative batches of 100 inputs, and a success will be considered only if, for none of the batches, the detector detects significant differences between the empirical distribution $\hat{\mathcal{P}}(Y)$ and $\mathcal{P}(Y)$.

The results are shown in Table \ref{tab:shift_detection}. As can be seen, our method succeeded in $24.8\%$ to $43.4\%$ percent of the cases depending on the configuration of the source distribution $P(Y)$, which is a reasonably high percentage considering the presence of a label-shift detection mechanism.  Furthermore, in the three cases a high fooling rate was maintained, of approximately 62\%, which supposes a loss of approximately 10\% in comparison to the maximum fooling rate that can be achieved in each case, which is shown in the fourth column. The fifth column of the table shows the average similarity between $\hat{\mathcal{P}}(Y)$ and $\widetilde{\mathcal{P}}(Y)$ according to the following metrics: the Kullback-Leibler divergence, the maximum absolute difference and the mean absolute difference. Only those cases for which the label-shift detector did not detect significant changes were considered. According to the three metrics, our method was capable of closely approximating the target distributions, achieving, for instance, an average Kullback-Leibler divergence of approximately 0.04 in the three cases.

\begin{figure}
\centering
\includegraphics[scale=0.4,valign=t]{{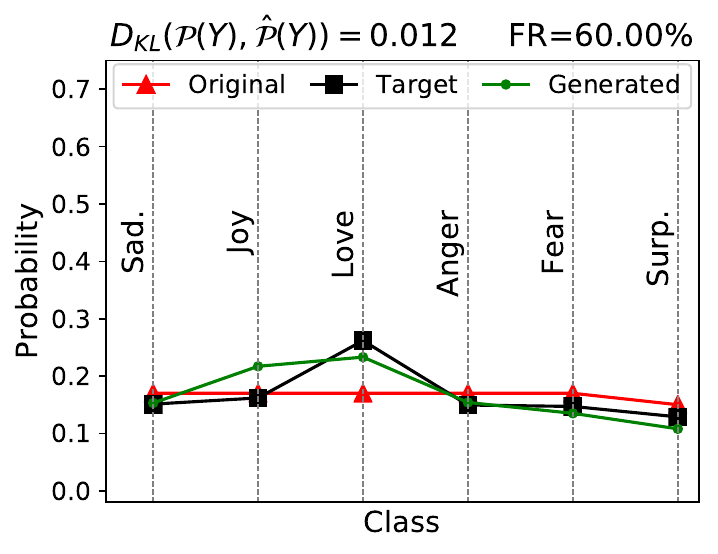}}
\includegraphics[scale=0.4,valign=t]{{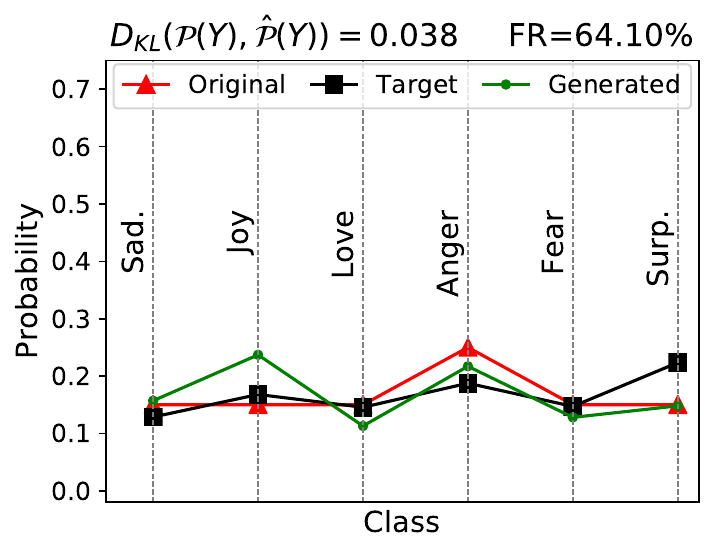}}
\includegraphics[scale=0.4,valign=t]{{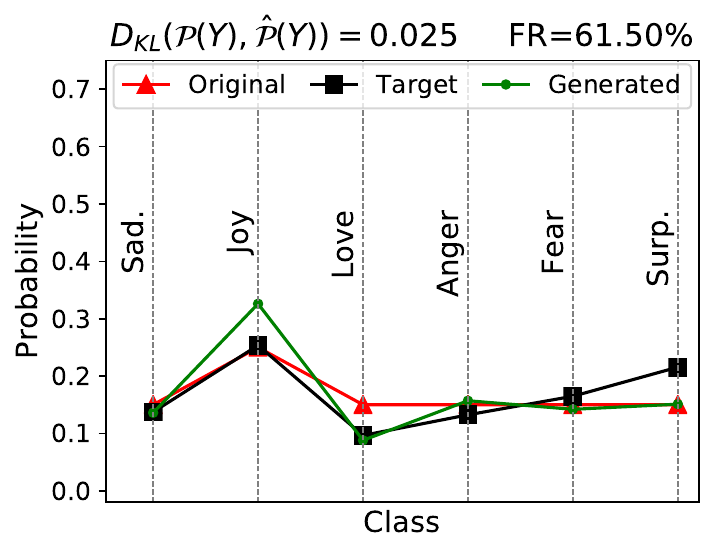}}
\includegraphics[scale=0.4,valign=t]{{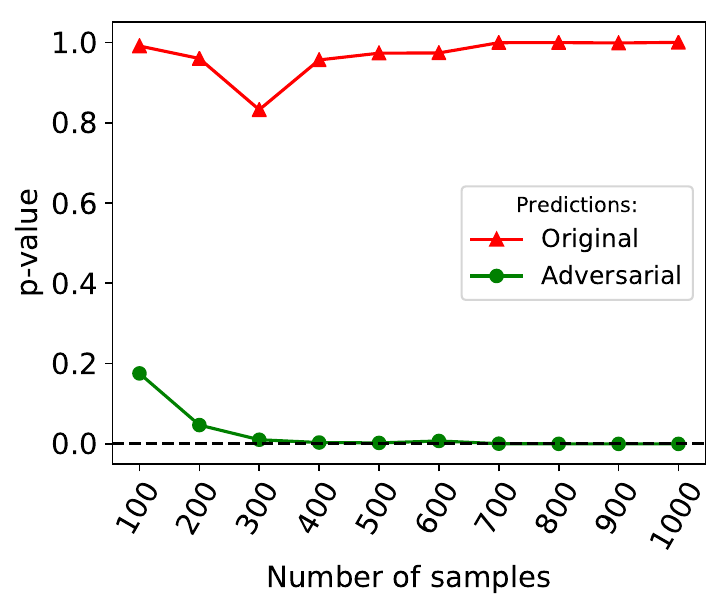}}
\includegraphics[scale=0.4,valign=t]{{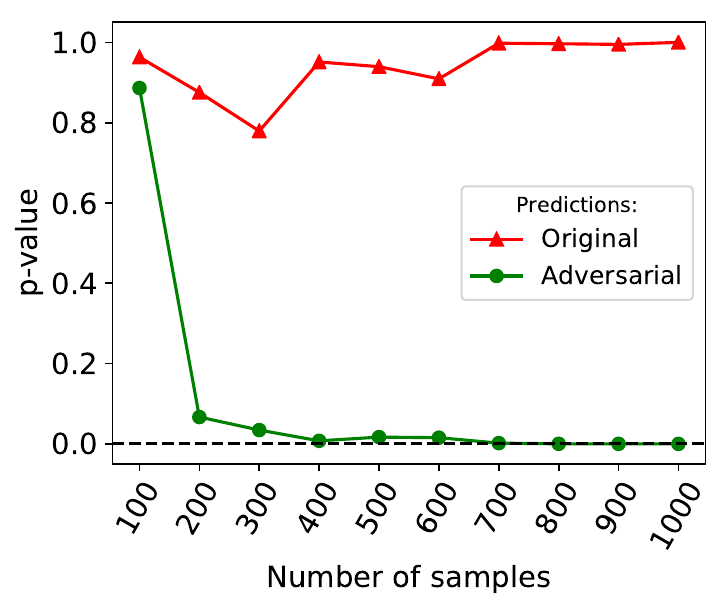}}
\includegraphics[scale=0.4,valign=t]{{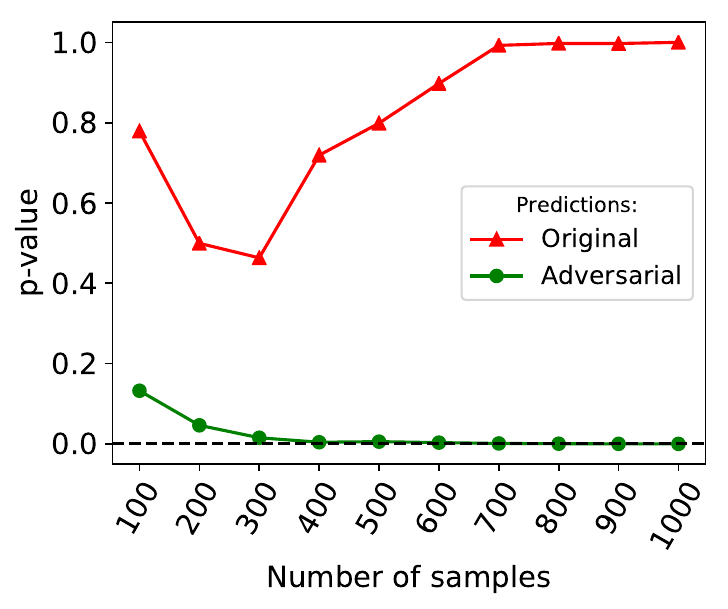}}
\caption{Illustrative label-shifts generated for the Tweet Emotion Classification task (column-wise). The first row provides a comparison of the source, target and produced probability distributions. In each case, the achieved fooling rate and Kullback-Leibler divergence between the target and the generated distributions is shown above the figure. The \linebreak second row shows, for each case, the evolution of the p-value computed by the BBSD label-shift detector during the attack process, evaluated in cumulative batches of 100 inputs and in both the correct predictions (i.e., when the model is not attacked) and the {adversarial}\linebreak predictions (i.e., when our method is applied). The dashed lines mark the detection threshold.} 
\label{fig:streaming_results}
\end{figure}

Finally, Figure \ref{fig:streaming_results} (top row) shows three illustrative label-shifts generated in our experiments, one for each of the source distributions considered (column-wise). The second row of the figure shows, for each case, the evolution in the p-value computed by the BBSD label-shift detector during the attack process, measured for cumulative batches of 100 inputs. For comparison, the p-value has been computed considering i) the adversarial predictions provided by the model when it is attacked, and ii) the original predictions, that is, the ones that would be provided if the model was not attacked.
As can be observed, both the target and empirical probability distribution of the classes represent an interpretation that can be considerably different from that of the original distribution. In the first case (left column), in which $\mathcal{P}(Y)$ initially portrays a \textit{uniform} opinion distribution in the population, the adversarially generated distribution portrays a predominantly positive opinion. A similar effect is achieved in the second case (middle column), in which the mode of the distribution is changed from a negative opinion to a positive opinion. Finally, in the third case (right column), the probability assigned to the mode of $\mathcal{P}(Y)$ is further increased (by reducing the probability assigned to some of the other classes), further biasing the distribution in favor of that mode.

\begin{table}[]
\centering
\scalebox{0.9}{
\begin{tabular}{@{}lccccc@{}}
\toprule
$\mathcal{P}(Y)$ & 
Success (\%) &
FR (\%) &
\begin{tabular}[c]{@{}c@{}}Max. \\ FR (\%)\end{tabular} &
\begin{tabular}[c]{@{}c@{}}Similarity\\ \ (KL \ / \ Max. / Mean)\end{tabular}
\\ \midrule
Uniform & 27.90 & 60.99 & 78.30 & 0.03 / 0.08 / 0.03\\
Tweak-2 & 43.40 & 62.63 & 77.20 & 0.05 / 0.10 / 0.04\\
Tweak-4 & 24.80 & 61.87 & 78.80 & 0.04 / 0.08 / 0.04\\
\bottomrule
\end{tabular}
}
\caption{Attack performance of the EWTM in producing label-shifts in the presence of the BBSD label-shift detection method. The following information is provided, column-wise: source distribution $\mathcal{P}(Y)$, percentage of cases in which the label-shift was not detected by the BBSD, average fooling rate achieved by the attacks, maximum fooling rate achievable (as reference), and the average similarity between the produced and the target probability distributions. The similarity is reported for three different metrics: Kullback-Leibler (KL) divergence, maximum absolute error and mean absolute error.}
\label{tab:shift_detection}
\end{table}

\section{Conclusions}
\label{sec:conclusions}
In this paper, we have introduced a novel strategy to generate adversarial attacks capable of producing not only prediction errors in machine learning models, but also any desired probability distribution for the classes when the attack is applied to multiple incoming inputs. This multiple-instance attack paradigm, due to its capability of coordinating multiple attacks to produce more complex malicious behaviors in the models, exposes threats that cannot be conducted by the conventional paradigms, broadening the horizon of adversarial attacks. The proposed attack methodology
has been conceived as an extension of targeted adversarial attacks, in which the target class is stochastically selected under the guidance of a transition matrix, which is optimized to achieve the desired goals. We have introduced four different strategies to optimize the transition matrices, which can be solved by using linear programs. Our approach was experimentally validated for the spoken command classification task, using different targeted adversarial attack algorithms as a basis. Furthermore, we also evaluated the success of our methods in preventing the attacks from being detected by label-shift detection methods in a streaming classification scenario.
Our results clearly show that the introduced methods are capable of producing close approximations of the target probability distribution for the output classes while achieving high fooling rates. 

As future research, the introduced approaches could be extended to generate adversarial class distributions using a single \textit{universal} perturbation. In this way, a single perturbation may not only cause the misclassification of every input, but also produce a desired probability distribution of the classes when applied to a large number of samples. Similarly, a generative adversarial network can be trained to produce both objectives at the same time. Apart from that, we plan to extend our methods to consider more challenging scenarios, such as highly imbalanced classification problems or scenarios where the source probability distribution of the classes changes over time, which is often the case in practice.

Finally, the introduced methods can be extended to generate different types of attacks. For instance, an adversary might be interested in approximating a target probability distribution of the classes while fooling the model the least possible times, which can be achieved by maximizing the values in the diagonal of the transition matrices. Similarly, by including simple restrictions in the linear programs used to optimize the transition matrices, the adversary can choose to fool the model more often for inputs of some classes than for others, decide not to fool inputs of some classes, or specify beforehand other kinds of transition patterns.

Overall, the study of such novel types of adversarial attacks contributes to exposing new vulnerabilities of current machine learning models, and previewing defenses for such weaknesses, which is essential for the development of a more reliable and ethical application of these models.\\

\acks{This work is supported by the Basque Government (KK2020/00049 project through ELKARTEK program, BERC 2022-2025 program, and PRE\_2019\_1\_0128 predoctoral grant) and by the Spanish Ministry of Science, Innovation and Universities (project PID2019-104966GB-I00 and FPU19/03231 predoctoral grant). Jose A. Lozano acknowledges support of the Spanish Ministry of Science, Innovation and Universities through BCAM Severo Ochoa accreditation (SEV-2017-0718).}

\appendix

\section{Overview of the Selected Adversarial Attacks}
\label{sec:supp_description_attacks}

In this section, we describe the adversarial attacks used to validate our methods.

\subsection{DeepFool}
\label{sec:supp_description_deepfool}
The DeepFool algorithm \citep{moosavi2016deepfool} consists of perturbing an initial input $x_0$ towards the closest decision boundary of the decision space represented by the model. Due to the intractability of computing these distances in high-dimensional spaces, a first-order approximation of the decision boundaries is employed, and the input is iteratively pushed towards the (estimated) closest decision boundary at each step until a wrong prediction is produced.  Precisely, being $f_j$ the output logits of a classifier $f$ corresponding to the class $y_j$, $f'_j=f_j(x_i') - f_{f(x_0)}(x_i')$ and  $w'_j=\bigtriangledown f_j(x_i') - \bigtriangledown f_{f(x_0)}(x_i')$, the following update-rule is employed:
\begin{equation}
\label{eq:deepfool_update_rule}
\displaystyle  x_{i+1}' \leftarrow x_i' + \frac{| f'_l |}{\ \ || w'_l ||_2^2}w'_l, \ \  \ l =
\argmin_{j\neq f(x_0)} \frac{ | f'_j |}{\ \ || w'_j ||_2},
\end{equation}
in which $w'_l$ represents the direction towards the (estimated) closest decision boundary, corresponding to the class $y_l$, and $\frac{| f'_l |}{\ \ || w'_l ||_2}$ the step size. The targeted version of DeepFool can be obtained if, at every iteration $i$, the sample is moved in the direction of the target class $y_t \! \neq \! f(x_0)$, that is:
\begin{equation}
\displaystyle  x_{i+1}' \leftarrow x_i' + \frac{|f'_t |}{\ \ || w'_t ||_2^2}w'_t.
\end{equation}
In this case, the process stops when the condition $f(x_i')=y_t$ is satisfied. The algorithm was restricted to a maximum of 30 iterations in our experiments.

\subsection{Gradient Based Approaches}
\label{sec:supp_description_gradient_based}
In \citet{goodfellow2014explaining}, a single-step gradient ascent approach was proposed, called \textit{Fast Gradient Sign Method} (FGSM), to efficiently generate adversarial perturbations. The attack strategy is based on linearizing the cross-entropy loss $L(x,y)$, where $y=f(x)$, and perturbing the input in the direction determined by the gradient of $L(x,y)$ with respect to the input $x$, $\nabla L_x(x,y)$. Thus, the adversarial example is generated according to the following closed formula: 
\begin{equation}
x' = x + \epsilon \cdot \text{sign} (\nabla L_x(x,y)),
\end{equation}
where $\text{sign}(\cdot)$ is the sign function and $\epsilon$ a budget parameter that controls the $\ell_\infty$ norm of the perturbation. The drawback of the FGSM is that a single step might not be enough to change the output class of the model. To solve this limitation, this strategy can be extended to iteratively perturb the input in the direction of the gradient:
\begin{equation}
x_{i+1}' = \mathcal{P}_{x,\epsilon}(x_i' + \alpha \cdot \text{sign} (\nabla L_x(x_i',y))),
\end{equation}
where $\alpha$ controls the step size and the projection operator $\mathcal{P}_{x,\epsilon}$ ensures that ${||x'-x||_\infty \leq\epsilon}$. This attack is known as the Projected Gradient Descent (PGD) \citep{madry2018deep}. In this paper, the PGD algorithm was restricted to a maximum of 30 iterations. In both cases, a targeted formulation can be obtained by considering the loss with respect to the target class $y_t$,  $L(x,y_t)$, and perturbing $x$ in the opposite direction of the gradients, that is, $\text{sign}(-\nabla L_x(x,y))$.

\subsection{Carlini and Wagner Attack}
\label{sec:supp_description_cw}
In \citet{carlini2017towards}, the problem of generating an adversarial example is formulated as the following optimization problem (hereinafter referred to as the C\&W attack):
\begin{equation}
\label{eq:cw_formulation}
\text{minimize} \ \ ||\frac{1}{2}(\text{tanh}(w)+1)-x||_2^2 + c \cdot \mathcal{L}\left(\frac{1}{2}(\text{tanh}(w)+1)\right).
\end{equation}
where $\mathcal{L}$ is the following loss function:
\begin{equation}
\label{eq:cw_loss}
\mathcal{L}(x) = \text{max}(\text{max}_{i\neq t}\{f_i(x)\}  - f_t(x) , -\kappa).
\end{equation}
The adversarial example is defined as $x'=\frac{1}{2}(\text{tanh}(w)+1)$, which allows an unconstrained variable $w$ to be optimized, while ensuring that each value of the adversarial input is in a valid range, typically $[0,1]$. The parameter $\kappa$ in Equation \eqref{eq:cw_loss} controls the desired confidence in the incorrect class $y_t$, and the constant $c$ in Equation \eqref{eq:cw_formulation} balances the trade-off between the perturbation norm and the confidence in the incorrect class. In this paper, $\kappa$ is set to $0$ and a binary search is used to tune the parameter $c$ for every input. The attack was restricted to a maximum of 1000 optimization steps.

\subsection{Overview}

The selected adversarial attacks employ different strategies to generate the adversarial perturbations and to restrict the amount of perturbation. In the DeepFool algorithm, the perturbation is constructed in a greedy fashion, and the norm of the perturbation is not subject to any constraint. In the case of the FGSM, the parameter $\epsilon$ determines the $\ell_\infty$ norm of the perturbation beforehand. Similarly, the maximum $\ell_\infty$ norm of the perturbation is explicitly specified beforehand in the PGD method. Finally, in the C\&W attack, the norm of the perturbation is modelled as a term to be minimized in the optimization problem. Therefore, the selected attacks will be used to illustrate the effectiveness and validity of our approaches for different underlying adversarial attack strategies.

\section{Results with Different Adversarial Attacks}
\label{sec:supp_results_other_attacks}

\begin{figure}[]
\centering
\includegraphics[scale=0.45]{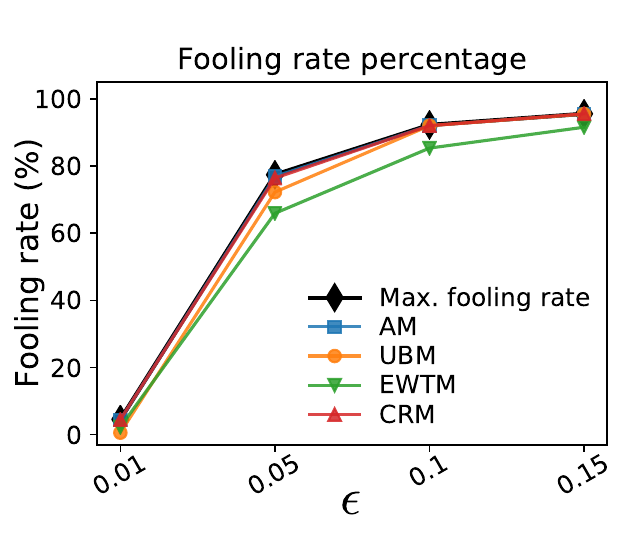}
\includegraphics[scale=0.45]{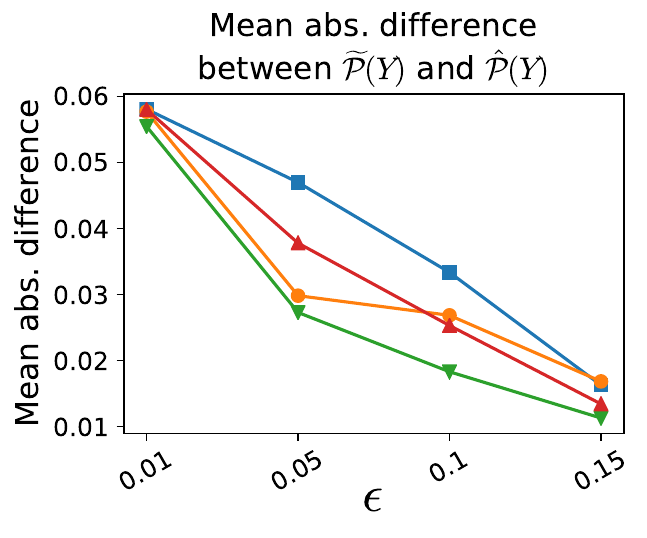}
\includegraphics[scale=0.45]{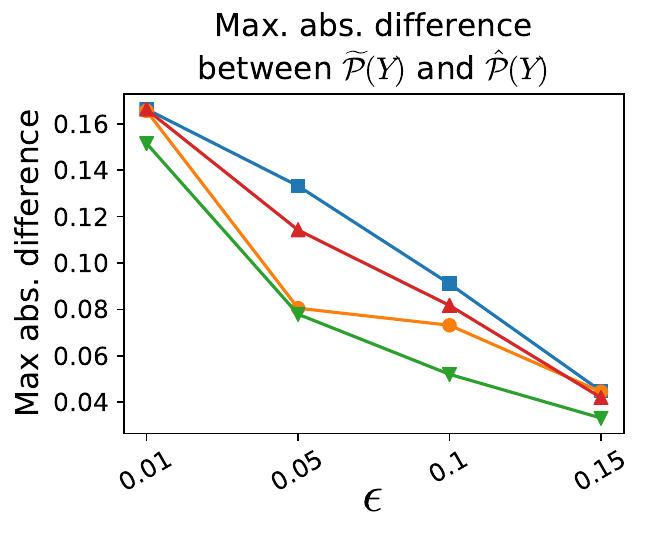}
\includegraphics[scale=0.45]{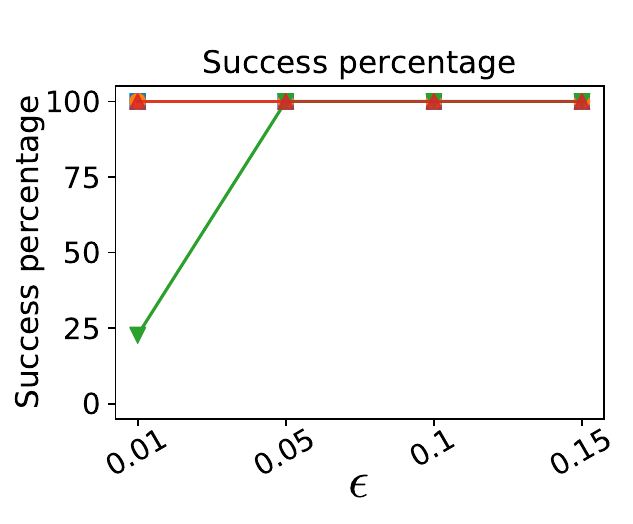}
\includegraphics[scale=0.45]{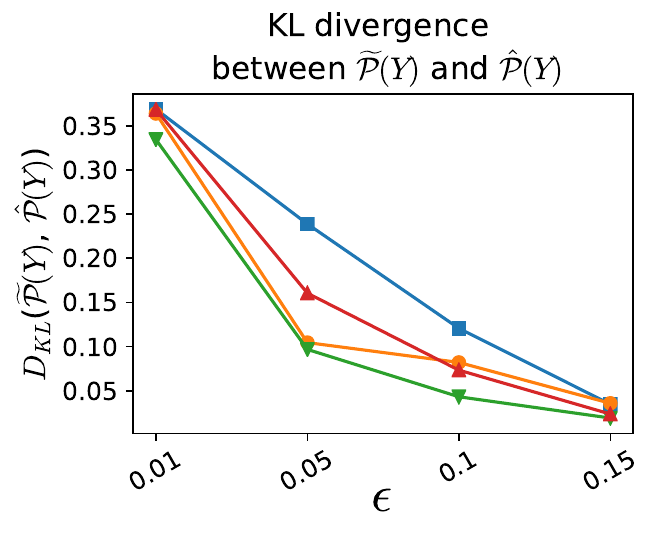}
\includegraphics[scale=0.45]{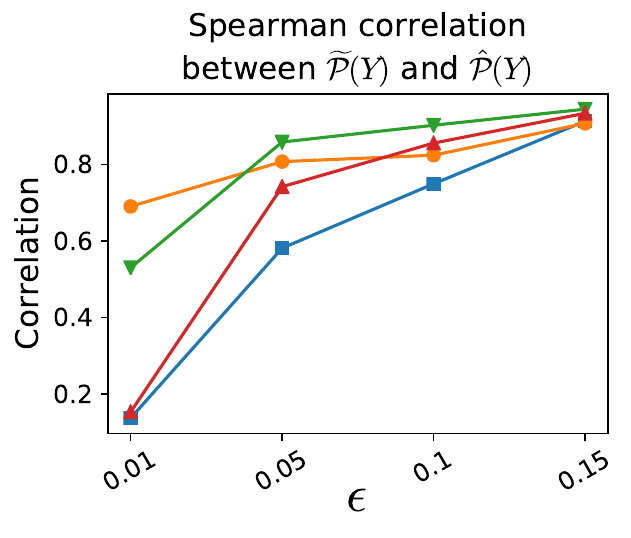}
\caption{Performance of the proposed methods using the Carlini \& Wagner adversarial attack.}
\label{fig:results_other_attacks_cw}
\end{figure}

\begin{figure}[]
\centering
\includegraphics[scale=0.45]{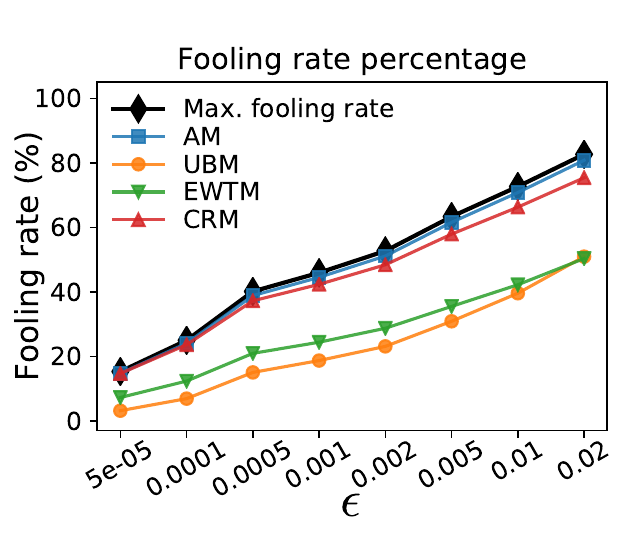}
\includegraphics[scale=0.45]{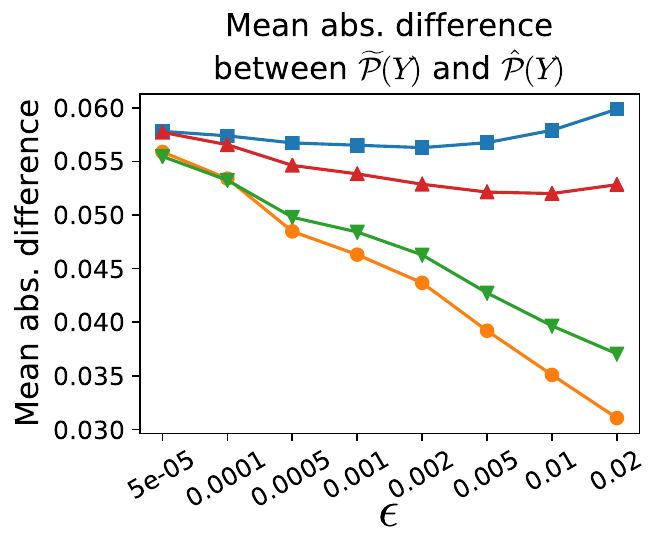}
\includegraphics[scale=0.45]{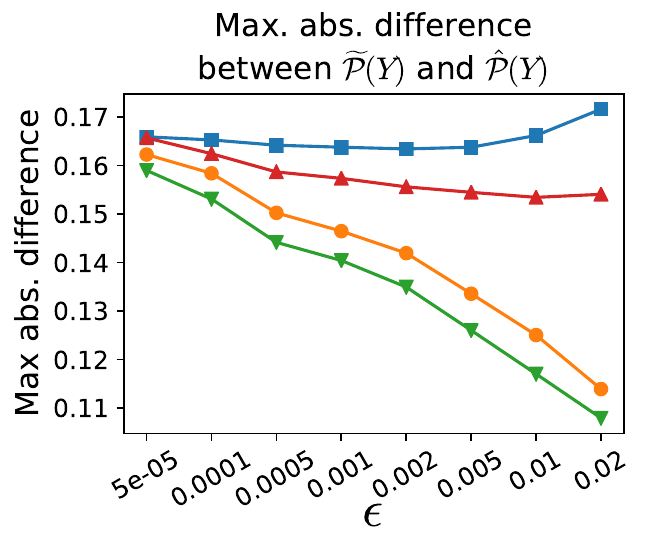}
\includegraphics[scale=0.45]{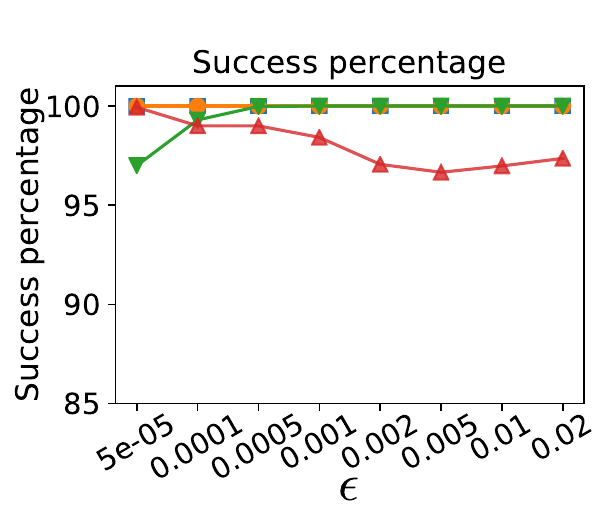}
\includegraphics[scale=0.45]{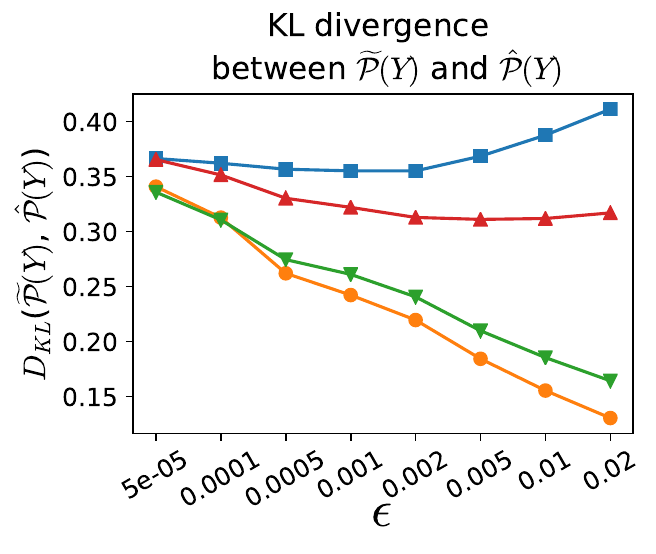}
\includegraphics[scale=0.45]{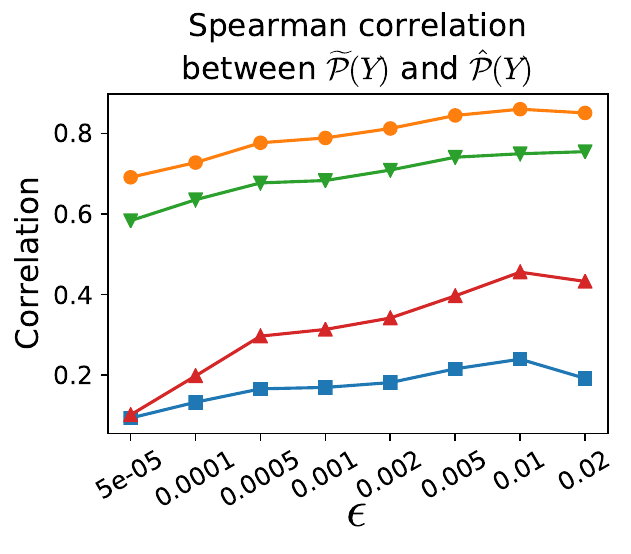}
\caption{Performance of the proposed methods using the Fast Gradient Sign Method.}
\label{fig:results_other_attacks_fgsm}
\end{figure}

\begin{figure}[]
\centering
\includegraphics[scale=0.45]{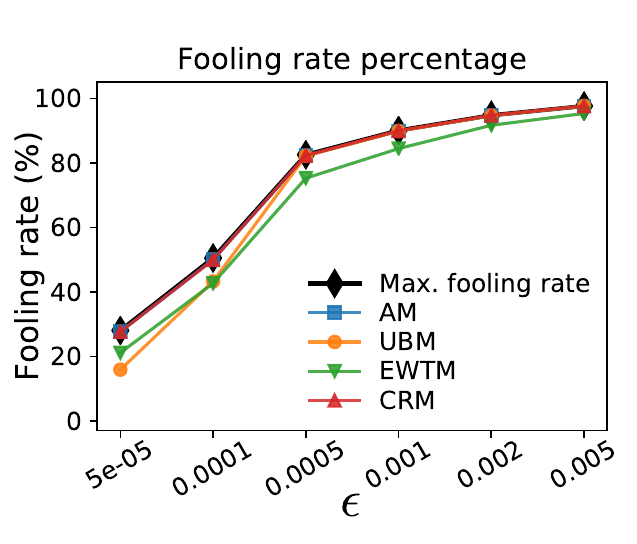}
\includegraphics[scale=0.45]{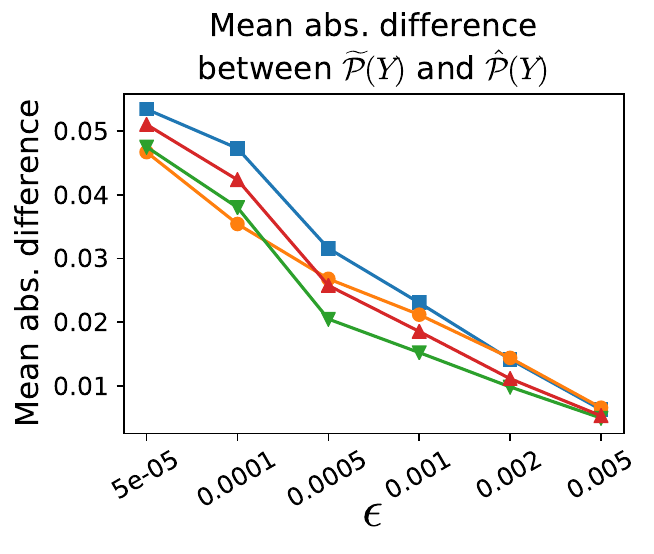}
\includegraphics[scale=0.45]{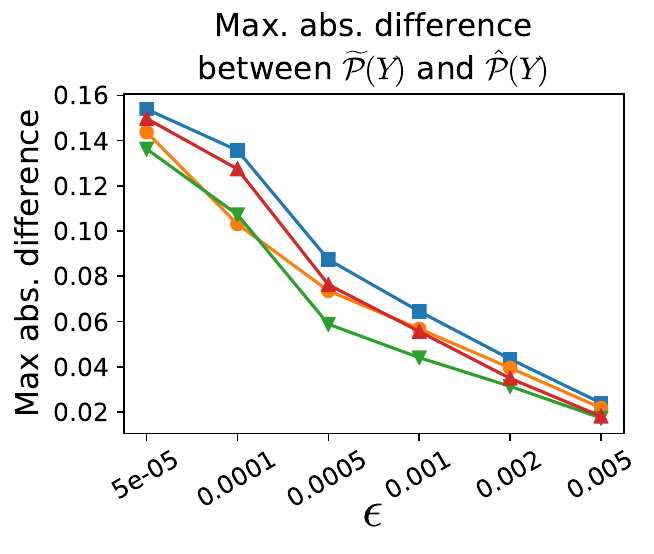}
\includegraphics[scale=0.45]{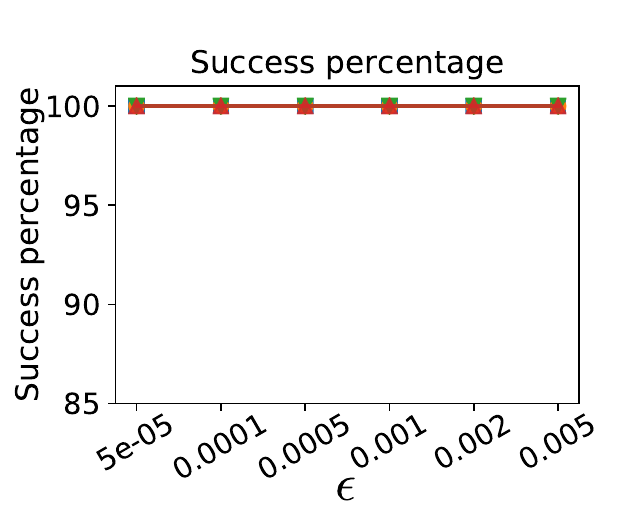}
\includegraphics[scale=0.45]{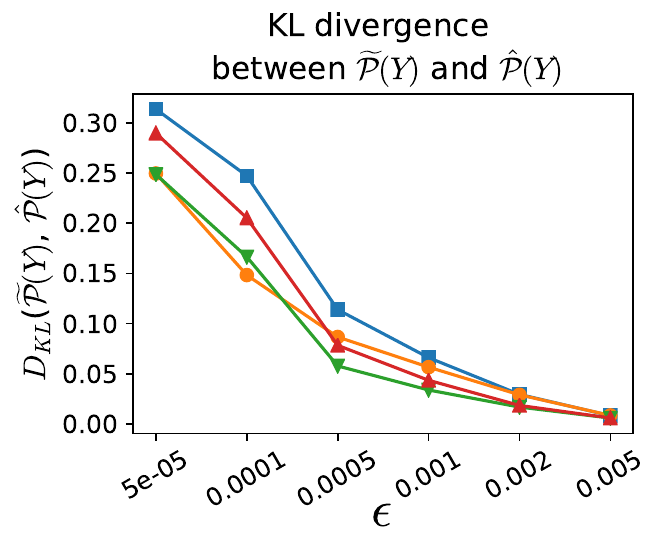}
\includegraphics[scale=0.45]{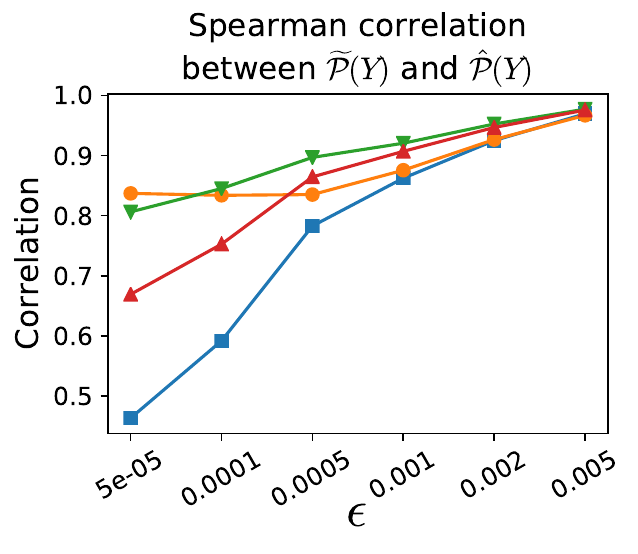}
\caption{Performance of the proposed methods using the Projected Gradient Descent attack.}
\label{fig:results_other_attacks_pgd}
\end{figure}

In this section we show that the methods proposed in this paper can be applied to a wide range of targeted attacks. For that purpose, the experimentation described in Section \ref{sec:results_general_case} is repeated, employing the following adversarial attacks: C\&W, FGSM and PGD. The results are shown in Figure \ref{fig:results_other_attacks_cw}, \ref{fig:results_other_attacks_fgsm} and \ref{fig:results_other_attacks_pgd}, respectively. The following performance metrics are shown for each attack: in the first row, fooling rate (left), mean absolute difference (center) and maximum absolute difference (right), and, in the second row, success percentage (left), Kullback-Leibler divergence (center) and Spearman correlation (right). The results are shown for different values of $\epsilon$, which have been selected to cover a representative range of fooling rate for each attack. To be consistent with the $\ell_p$ norm used to generate the perturbations in the FGSM and PGD methods, the $\ell_\infty$ norm has been limited for these attacks instead of the $\ell_2$ norm.

\begin{figure}[]
\centering
\includegraphics[scale=0.47]{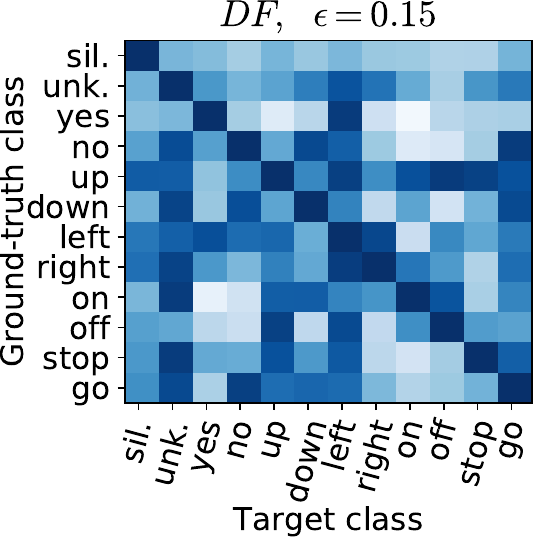}
\hspace{-0.7em}
\includegraphics[scale=0.47]{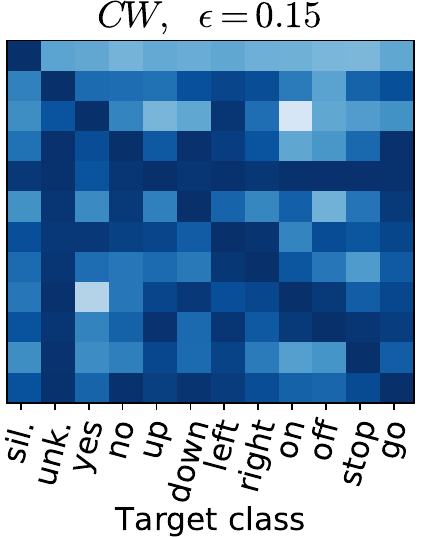}
\hspace{-0.7em}
\includegraphics[scale=0.47]{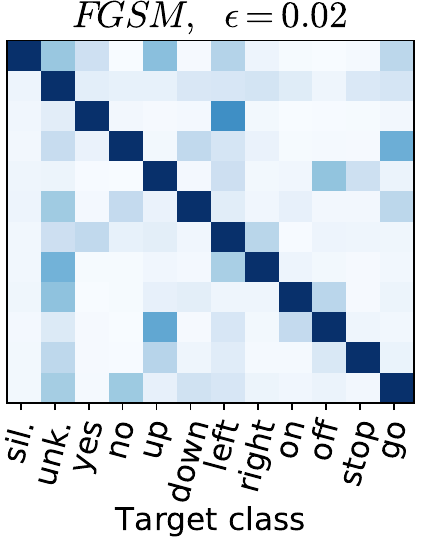}
\hspace{-0.7em}
\includegraphics[scale=0.47]{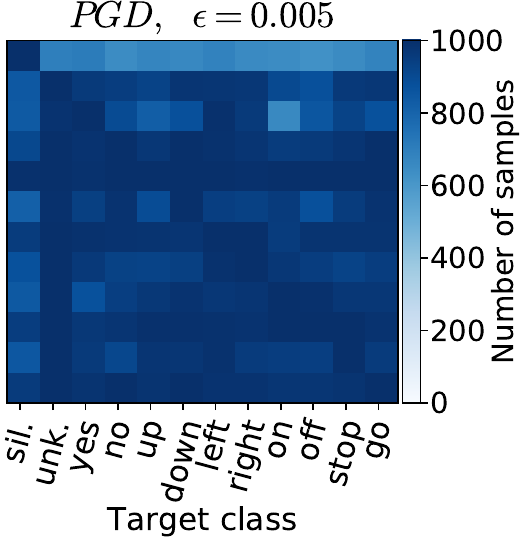}

\caption{R matrices (see Equation \ref{eq:matrix_r}) obtained with different attack strategies: DeepFool (DF), Carlini \& Wagner attack (C\&W), Fast Gradient Sign Method (FGSM), and Projected Gradient Descent (PGD). These results have been computed using the maximum value of $\epsilon$ evaluated in the experiments for each attack.}
\label{fig:R_total}
\end{figure}

Independently of the underlying adversarial attack employed, the results are comparable to those reported in Section \ref{sec:results_general_case}. Comparing the overall effectiveness obtained with each method, in all the cases the EWTM achieved the best results in approximating the target distributions, the AM the worst results, and the UBM and the CRM intermediate results. On the other hand, the EWTM achieved lower fooling rates in comparison to the other methods, which achieved values close to the optimal fooling rate. Nonetheless, more general conclusions can also be drawn by analyzing the results according to different factors, such as the reach of the underlying attack, that is, the number of samples that can be moved from one class to another without exceeding the norm restrictions. To better assess this factor, Figure \ref{fig:R_total} shows, for each attack, the frequency of each class transition in the set of samples $\bar{\mathcal{X}}$ considered in our experiments.\footnote{It is important to note that the attack strategies and restrictions considered have been selected to illustrate the effectiveness of our approaches in different scenarios, precisely, when the capabilities of the underlying adversarial attack is limited according to different factors, such as the maximum distortion allowed or the number of steps (that is, sacrificing effectiveness for efficiency). Therefore, these results should be taken as a comparison of the four methods introduced in this paper, rather than an exhaustive or representative comparison between the effectiveness of the adversarial attacks, as the restrictions or parameters set to each attack are not necessarily comparable or equivalent (for example, increasing the number of iterations for DeepFool would increase its reach).} These results have been computed for the maximum value of $\epsilon$ considered for each attack.

As can be seen, the greater the reach to the incorrect classes and the more regular this reach is among the possible pairs of source-target classes (which occurs for both C\&W and PGD attacks), the greater the similarity between the performance of the four methods. Moreover, even the AM achieved a high effectiveness in such scenarios, although in all the cases the remaining methods achieved a superior performance. In contrast, when the reach is considerably more irregular and sparse, as occurs for the FGSM, the differences between these methods is more pronounced. These results corroborate the finding that the strategies employed in the UBM, the EWTM and the CRM are capable of taking advantage of the information about the problem to better approximate the target distributions, especially in the more challenging scenarios in which few class transitions can be produced. Moreover, for the FGSM attack, the UBM achieved the best results in approximating the target distributions, being the method that best adapted to the challenging scenario imposed by the low reach of that attack, yet at the expense of obtaining lower fooling rates.
Finally, the success percentage of the CRM slightly decreased when the FGSM was employed, which might reveal that this method is not completely effective when the reach is very sparse, although the success percentage was above $96$\% in all the cases.

Overall, these results corroborate the conclusions reported in the paper, and show the validity of our approaches to effectively guide a wide range of adversarial attacks.

\newpage

\section{Reducing the Size of the Set $\mathcal{X}$}
\label{sec:supp_less_train}

In this section, we analyze the effect of reducing the number of samples per class $N$ (i.e., the total number of samples is $12N$, which is the size of the set $\mathcal{X}$) that are used to generate the transition matrices on the effectiveness of the methods. For this purpose, we repeated the experiments described in Section \ref{sec:results_general_case} using different values for $N$: $\{1,10,50,100,500\}$. As the set $\bar{\mathcal{X}}$ used in the $k$-fold cross-validation to validate our methods is composed of $1000$ samples per class, the number of folds will be determined by $k=\frac{1000}{N}$.  The cross-validation was repeated 50 times when $N=500$, 10 times when $N=100$, 5 times when $N=50$ and a single time when $N=10$ and $N=1$. Finally, the results will be computed using a maximum distortion threshold of $\epsilon=0.01$, and the analysis will focus on the DeepFool algorithm and on the AM, the UBM and the EWTM.

Regarding the percentage of success in finding feasible solutions to the linear programs, both the AM and the UBM maintained a success rate of 100\% for all the values of $N$ tried. For the EWTM, a success rate of 100\% was obtained when $N\geq 50$, 84.6\% when $N=10$ and 0.8\% when $N=1$, which shows that this method is not capable of producing valid transition matrices when a low number of samples per class is available.\footnote{As a single k-fold cross-validation is performed for $N=10$ and $N=1$, the success percentage has been computed as the number of folds in which a valid transition matrix is obtained, and this value has been averaged for the $100$ target probability distributions considered.}

\begin{figure}[]
\centering
\hspace{-0.4cm}
\includegraphics[scale=0.57]{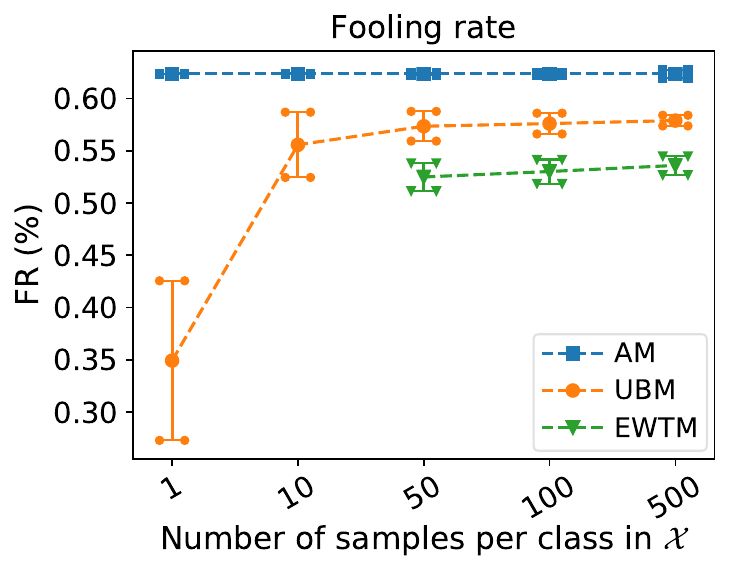}
\includegraphics[scale=0.57]{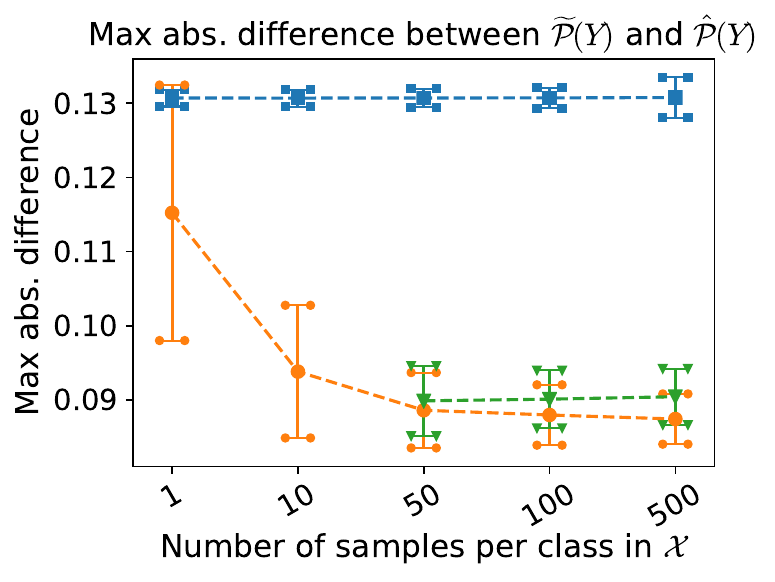}
\includegraphics[scale=0.57]{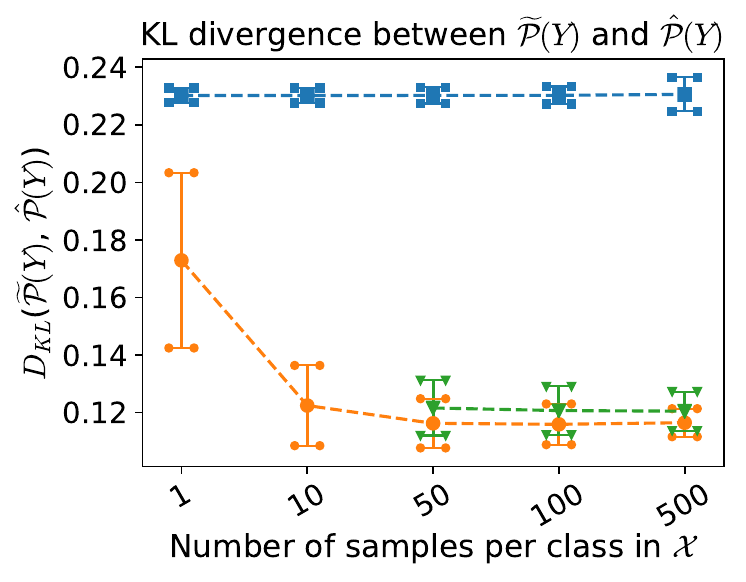} \ 
\includegraphics[scale=0.57]{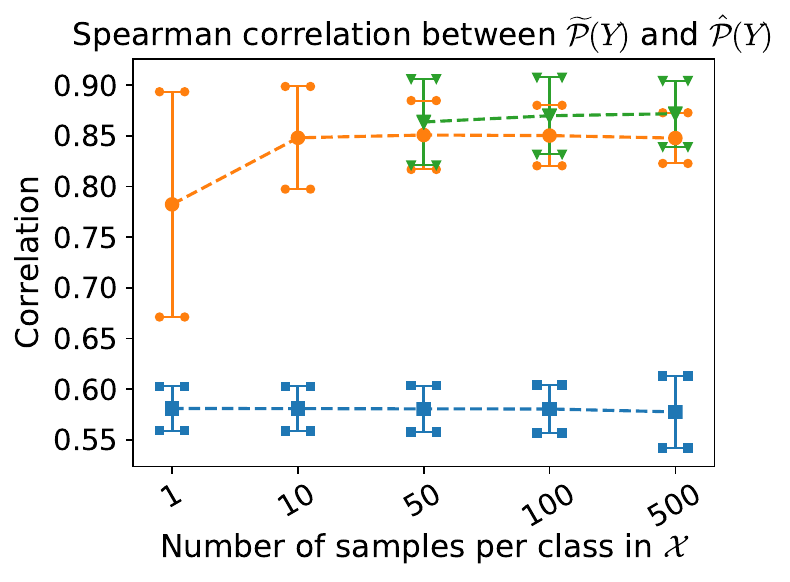}
\caption{Effectiveness of the introduced methods for different numbers of samples per class ($N$) used to generate the transition matrices: $N=\{1,10,50,100,500\}$. The results are shown for the AM, the UBM and the EWTM, considering the following metrics: Fooling rate (top-left), maximum absolute difference (top-right), Kullback-Leibler divergence  (bottom-left) and Spearman correlation (bottom-right). In each figure, for each value of $N$, the standard deviation has been included, represented using vertical bars.}
\label{fig:reduced_train}
\end{figure}

Regarding the effectiveness of the methods in approximating the target distributions, the following performance metrics are shown in Figure \ref{fig:reduced_train}, independently for each method: fooling rate (top left), maximum absolute difference (top right), Kullback-Leibler divergence (bottom left) and Spearman correlation (bottom right). In every figure, for each value of $N$, the average result obtained for the different cross-validations is shown, as well as the average standard deviation obtained for each target distribution along all the folds evaluated, which is depicted by vertical bars. Only the cases in which a success rate of 100\% is achieved by the methods are shown, thus ommiting the results corresponding to the EWTM when $N\leq 10$. According to the results, although the effectiveness decreases when $N$ is highly reduced (e.g., $N\leq 10$),  a high effectiveness is maintained even when the number of samples per class is reduced to $N=50$. These results show that a considerably small number of inputs per class can be used to efficiently generate our attacks, and, also, that our attacks are effective (in the \textit{prediction phase}) even when they are applied to a number of samples considerably larger than the number of elements used to optimize the attacks.\footnote{The loss in the effectiveness when $N$ is reduced might also depend on the regularity with which inputs belonging to the same class can be sent to the remaining classes, and, therefore, the loss could be higher in those problems in which there exists a low regularity.}

\section{Implementation Details}
\label{app:reproducibility}

Our code is publicly available at: \url{https://github.com/vadel/ACPD}. In what follows, we briefly describe how the results of this paper can be reconstructed using our repository. 
The required Python packages can be consulted in the \verb+setup/+ directory. 
This directory also contains an executable script that can be used to download the required datasets, models, and other additional resources, as well as to create the required directory structure. 

The main experimental pipeline 
is organized into two different parts: the generation of adversarial examples and the generation of adversarial class probability distributions. In the first part, the adversarial examples are precomputed for the sake of efficiency, so that, afterward, the methods proposed to generate the adversarial class probability distributions can be evaluated without the need of repeating the same computation several times. The code required to generate adversarial examples can be found in the \verb+adv_attacks/+ directory, which contains two subdirectories, one for each of the two problems considered: the speech command classification problem employed in Section \ref{sec:validation} (\verb+speech_commands/+) and the Tweet emotion classification problem considered in Section \ref{sec:exp_shift_detection} (\verb+text/+). In the second part, the effectiveness of the methods introduced in this paper is evaluated. The code corresponding to this part is located in the \verb+optimization/+ directory.
The implementation of the methods introduced in Section \ref{sec:methods} and the baselines described in Section \ref{sec:experimental_details} can be found in the file \verb+acpd_methods.py+, in the same directory.

In order to reproduce the experiments reported in Section \ref{sec:validation}, first, the adversarial examples should be generated using the script \verb+deepfool/targeted_launcher.py+, which will compute, for each input, an adversarial example targeting each of the classes in the problem, using the DeepFool algorithm. To use attacks other than DeepFool (see Appendix \ref{sec:supp_results_other_attacks}), the script \verb+foolbox/targeted_launcher.py+ can be used instead, which makes use of the Foolbox package \citep{rauber2018foolbox}.  Once all the results corresponding to an adversarial 
\linebreak
attack have been obtained, they can be collected using the script 
\linebreak 
\verb+analysis/pack_results_targeted.py+. 
These collected results will be used to know which targeted attacks are feasible given a specific attack budget. 
In the second part, to evaluate the effectiveness of our methods, the script \verb+launcher_acpd_experiments.py+ can be employed, by specifying i) the attack method to be used, ii) the attack budget, and iii) the set of target distributions: either a single uniform distribution (Section \ref{sec:results_particular_case}) or a set of 100 random Dirichlet distributions (Section \ref{sec:results_general_case}).
In order to compute and visualize the average results of the experiments, the following Jupyter notebook (located in the \verb+visualization/+ directory) can be used: \verb+Visualize_General_Comparison.ipynb+. To visualize the source, target, and generated distributions for one particular case (as in Figure \ref{fig:targeting_initial}), the following notebook can be used instead: \verb+Visualize_Individual_Results.ipynb+.

Regarding the experiments carried out in Section \ref{sec:exp_shift_detection}, the adversarial examples will be generated using the same methodology described above, but, in this case, by means of the 
\linebreak
OpenAttack package \citep{zeng2021openattack} and the script \verb+openattack/launcher_attacks.py+.
Afterward, the main experiment can be executed using the notebook
\linebreak
\verb+Label_Shift_Evaluation_TweetClassification.ipynb+, and the final results can be summarized using the notebook \verb+Label_Shift_EvaluationSummary.ipynb+, both located in the \verb+label_shift/+ directory.

We refer the reader to our repository for a more extensive documentation, in which more particular details can be consulted.

\newpage

\bibliography{references}

\end{document}